%% file: main.tex
\documentclass[10pt,twocolumn,letterpaper]{article}

\usepackage[pagenumbers]{iccv} %

\input{preamble}
\definecolor{iccvblue}{rgb}{0.21,0.49,0.74}
\usepackage[pagebackref,breaklinks,colorlinks,allcolors=iccvblue]{hyperref}
\usepackage{orcidlink}

\def\paperID{***} %
\def\confName{****}
\def\confYear{2025}

\title{ORXE: Orchestrating Experts for Dynamically Configurable Efficiency}

\author{Qingyuan Wang\orcidlink{0000-0002-7879-4328}\quad Guoxin Wang\orcidlink{0000-0002-3619-3399}\quad Barry Cardiff\orcidlink{0000-0003-1303-8115}\quad Deepu John\orcidlink{0000-0002-6139-1100}\vspace{8pt}\\
University College Dublin, Ireland
}

\begin{document}

\twocolumn[{%
\renewcommand\twocolumn[1][]{#1}%
\maketitle
\begin{center}
    \centering
    \captionsetup{type=figure}
    \includegraphics[width=0.75\textwidth]{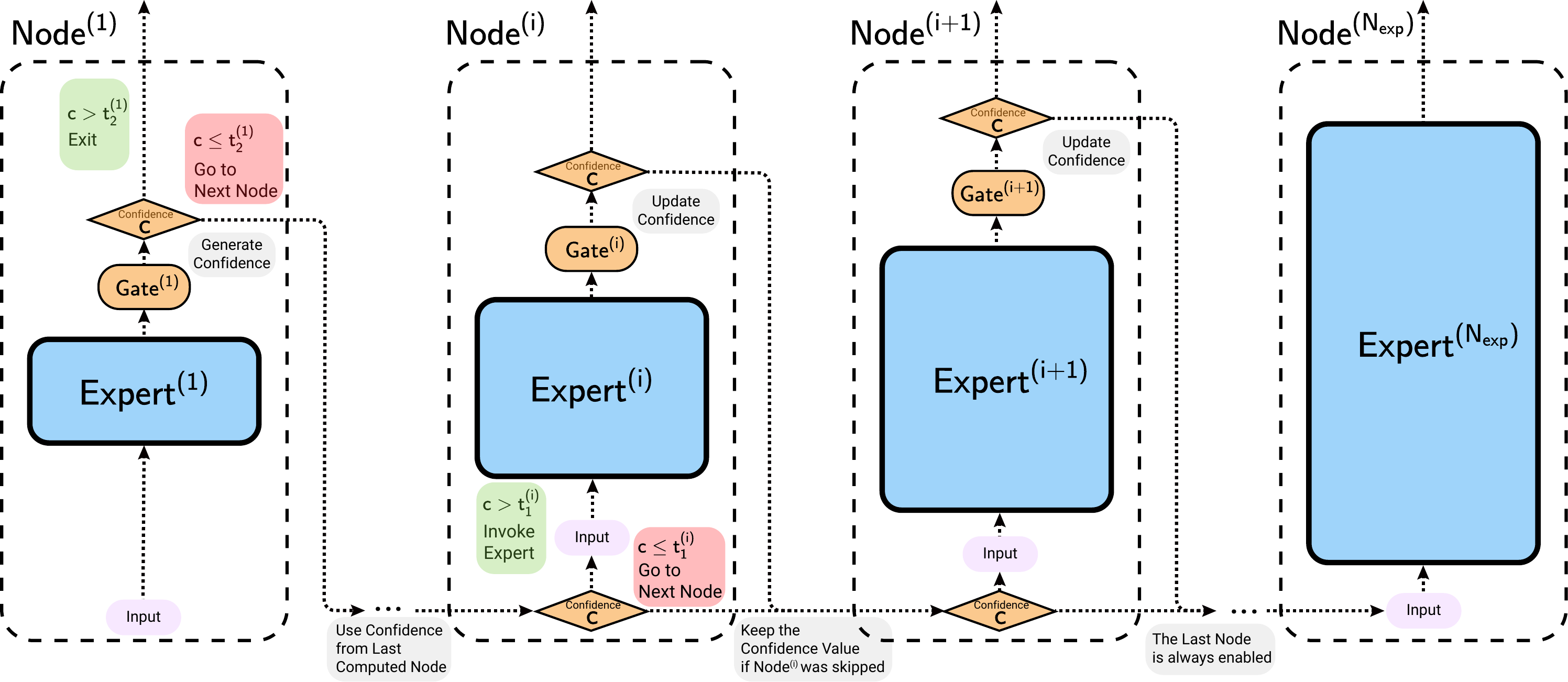}
    \captionof{figure}{The overview of proposed method. The system employs multiple pre-trained experts with incremental computational costs and performance levels. The input samples are dynamically routed through the experts based on its difficulty to experts. The overall system performance is adjustable by changing the threshold of each gate. $\mathit{Expert}^{(i)}$ and $\mathit{Gate}^{(i)}$ are the expert and the post-expert gate for the $i$-th node. $N_{\mathit{exp}}$ is the number of experts. The procedure starts with the first enabled node which produce the initial prediction and confidence. If the confidence is greater than the threshold $t_{2}^{(1)}$, the computation exits. Otherwise, it goes to the next enabled node. At the entry of an intermediate node, this node is skipped without any computation if the confidence from the previous node is too low.}
    \label{fig:overview}
\end{center}%
}]

\input{sec/0_abstract}    
\input{sec/1_intro}

\input{sec/2_method}

\input{sec/3_expriment}
{
    \small
    \bibliographystyle{ieeenat_fullname}
    \bibliography{doclib}
}
\input{sec/X_suppl}

\end{document}

%% file: preamble.tex
\usepackage[vlined ,ruled,linesnumbered]{algorithm2e}
\usepackage{graphicx}
\usepackage[normalem]{ulem}
\useunder{\uline}{\ul}{}

%% file: sec/0_abstract.tex
\begin{abstract}
This paper presents ORXE, a modular and adaptable framework for achieving real-time configurable efficiency in AI models. By leveraging a collection of pre-trained experts with diverse computational costs and performance levels, ORXE dynamically adjusts inference pathways based on the complexity of input samples. Unlike conventional approaches that require complex metamodel training, ORXE achieves high efficiency and flexibility without complicating the development process. The proposed system utilizes a confidence-based gating mechanism to allocate appropriate computational resources for each input. ORXE also supports adjustments to the preference between inference cost and prediction performance across a wide range during runtime. We implemented a training-free ORXE system for image classification tasks, evaluating its efficiency and accuracy across various devices. The results demonstrate that ORXE achieves superior performance compared to individual experts and other dynamic models in most cases. This approach can be extended to other applications, providing a scalable solution for diverse real-world deployment scenarios.
\end{abstract}

%% file: sec/1_intro.tex
\section{Introduction}
\label{sec:intro}
The development of AI applications is resource-intensive and typically produces a single fixed model, optimized for a specific environment. However, real-world environments are unpredictable and dynamic, with varying demands on resources and performance. For example, an online AI service might face a sudden surge in users, causing disruptions. An autonomous vehicle might encounter complex traffic that requires increased performance. Conventional fixed and monolithic models cannot dynamically adjust efficiency or performance, making them unsuitable for such variable scenarios.\\
Some methods\cite{cai_once-for-all_2020-1,yang_neural_2024,devvrit_matformer_2023,yu_slimmable_2018,wang_dyce_2024} have been proposed to support configurable efficiency for adapting the varying environment. These approaches often involve constructing a metamodel capable of generating submodels of different sizes. However, enabling configurable efficiency adds significant complexity, burdening the design and training process and preventing their performance from matching that of state-of-the-art models.\\
To address this challenge of adapting to varying environments while minimizing system reconstruction costs, we propose ORXE. ORXE leverages multiple pre-trained models, orchestrating them into a metamodel with an early exit mechanism\cite{teerapittayanon_branchynet_2017, wang_not_2021}. The overall inference cost and performance of the metamodel can be adjusted dynamically by modifying early exit configurations. As shown in \cref{fig:overview}, ORXE uses multiple expert models, ordered by size. Each expert, except the last, is enclosed by gates that generate confidence scores to determine if the current expert is sufficient for a given input. If not, the sample proceeds to the next expert for a better prediction.\\
The pre-expert gate rapidly assesses the sample. If it deems the sample too challenging, it skips the current node directly. Otherwise, the expert processes the sample. After processing, the post-expert gate evaluates the prediction. The post-expert gate is more accurate since it has access to the actual prediction. If this gate also returns a positive decision, the inference process for the given sample is concluded.\\
Unlike conventional compression methods like pruning\cite{blalock_what_2020, fang_depgraph_2023, li_pruning_2017}, quantization\cite{rokh_comprehensive_2023}, and distillation\cite{hinton_distilling_2015}, early exit reduces computational cost by assessing sample difficulty. Most real-life samples are relatively easy, allowing smaller models to process them, while more challenging cases are handled by larger models to maintain prediction quality. However, early exit methods often require specialized model design, involving exit branches attached to intermediate layers, which can disrupt training and limit performance. Some studies\cite{wang_tiny_2025} suggest that using independent models for exits may be more effective. The proposed method leverages multiple pre-trained models at different scales, reducing system construction costs. These models are independent from the gating mechanism, allowing development to focus solely on the task without added complexities.\\
In constructing the ORXE metamodel, we propose a search algorithm to generate configurations for different scenarios. The objective function incorporates a performance factor, $\lambda \in [0,1]$, to balance efficiency and performance. By iterating the search process with different $\lambda$ values, the algorithm yields optimal thresholds for various preferences. This configuration search can also be viewed as a form of network architecture search\cite{elsken_neural_2019-1,wu_training-free_2024}, applied post-training. During runtime, the ORXE metamodel can load any generated configuration to adapt to changing conditions.\\
Compared to the related approaches discussed earlier, ORXE incorporates many advantageous features and combines them in a streamlined manner. Specifically, the contributions of this work can be summarized by the following key features:
\begin{itemize}
    \item \textbf{Sample-Aware Compression:} The ORXE metamodel reduces inference costs for simple samples while maintaining performance for complex samples. This type of compression is distinct from conventional methods like quantization, pruning, and distillation, as it dynamically adapts to sample complexity.
    \item \textbf{Runtime Configurable Efficiency:} The proposed method allows for runtime adjustment of the efficiency-performance trade-off to precisely meet varying demands. Importantly, ORXE achieves this without compromising overall performance or adding design complexity.
    \item \textbf{Upgradable and Low Construction Cost:} Among methods that offer the aforementioned features, ORXE stands out for its flexibility and low construction cost. By decoupling these features from a monolithic design, ORXE can leverage existing models optimized for the original task, making it easily upgradable when new expert models become available.
    \item    \textbf{Flexible Deployment:}  The modular design of ORXE provides deployment flexibility, with each node of the metamodel capable of being deployed on different devices. Simple samples can be processed locally on edge devices, reducing the need for centralized resources. It also supports speculative execution, allowing subsequent nodes to compute concurrently with earlier nodes, thereby further reducing latency.
\end{itemize}

\section{Related Work}
\paragraph{Evolving Efficient Models} As AI models continue to evolve, there has been an ongoing effort to enhance efficiency in model development. Recent advancements in efficiency have been largely driven by innovations in network architecture search\cite{elsken_neural_2019-1, howard_searching_2019, qin_mobilenetv4_2024, tan_efficientnet_2019-1}, large-scale pre-training\cite{he_masked_2022, woo_convnext_2023-1,brown_language_2020} or architectural improvements\cite{he_deep_2016,vaswani_attention_2017,dosovitskiy_image_2021}. These approaches fundamentally enhance model capabilities, enabling new models to achieve superior performance under equivalent computational constraints. The research community has made substantial progress in this direction, resulting in the development of numerous innovative backbone architectures\cite{cai_efficientvit_2023, dai_coatnet_2021, ding_davit_2022, el-nouby_xcit_2021, li_next-vit_2022, liu_efficientvit_2023, liu_swin_2021, liu_swin_2022, soldaini_cascade_2020, tu_maxvit_2022, wu_tinyvit_2022, yu_metaformer_2023}. Following these foundational improvements, various general compression techniques, such as pruning\cite{blalock_what_2020, fang_depgraph_2023,li_pruning_2017}, quantization\cite{rokh_comprehensive_2023} and distillation\cite{hinton_distilling_2015}, have been applied to further reduce model complexity. The standard practice for model deployment involves selecting an appropriately scaled model and experimenting with different combinations of compression techniques. However, this pipeline often yields immutable monolithic models with fixed computational costs and performance characteristics, exclusively optimized for specific scenarios. Changes in deployment environments or user demands may necessitate redesign, re-optimization, or even retraining of these models, highlighting significant limitations in adaptability.

\paragraph{Dynamic Model and Sample-aware Compression} Model performance often correlates with model size\cite{alabdulmohsin_revisiting_2024, kaplan_scaling_2020, krajewski_scaling_2024}. Expanding the model size in regular static models leads to a substantial increase in computational requirements. In contrast, some dynamic models, such as the Sparsely Gated Mixture-of-Experts\cite{daxberger_mobile_2023, fedus_switch_2022, krajewski_scaling_2024, riquelme_scaling_2021, shazeer_outrageously_2017, gururangan_scaling_2023, komatsuzaki_sparse_2023, mustafa_multimodal_2022}, activate only a subset of the computation graph based on the input, which prevents computation from growing proportionally with the number of parameters. Additionally, other dynamic models can assess the difficulty of input samples and allocate resources accordingly. Techniques like Early Exit\cite{teerapittayanon_branchynet_2017,kaya_shallow-deep_2019,panda_conditional_2016,yang_resolution_2020,hang_msnet_2023,hong_panda_2021,chen_cf-vit_2022,han_dynamic_2023,wang_not_2021,han_learning_2022,wang_glance_2020,zhang_basisnet_2021} terminate computation early for simpler inputs. Layer Skipping\cite{wang_skipnet_2018,raposo_mixture--depths_2024, wu_blockdrop_2018,veit_convolutional_2020}bypasses intermediate layers when they are unnecessary and Dynamic Pruning\cite{gao_dynamic_2018, lin_runtime_2017-1,hua_channel_2019} selectively removes channels or neurons based on the input. These approaches offer an alternative route to achieving higher efficiency. However, dynamic models are inherently more challenging to design and train, as the network must adapt to changing structures.

\paragraph{Configurable Efficiency}
Some models are designed to be configurable after the training phase. These methods typically train a metamodel from which submodels are sampled to support various applications. OFA\cite{cai_once-for-all_2020-1} introduces a model that can be tailored and fine-tuned to fixed-sized smaller models. NeuMeta\cite{yang_neural_2024} learns implicit neural representation to predict variable-sized weights of submodels. Additionally, some approaches allow for efficiency preferences to be adjusted on the fly during inference.  For instance, layer sizes can vary in models like MatFormer\cite{devvrit_matformer_2023} and Slimmable Neural Networks\cite{yu_slimmable_2018} depending on the application. DyCE\cite{wang_dyce_2024} adjusts network depth by providing configurations for a multi-exit network. However, the design and training of metamodels are often significantly more complex, limiting their performance compared to state-of-the-art standalone models TinySaver\cite{wang_tiny_2025} explores the potential of utilizing pre-trained state-of-the-art models to implement configurable efficiency.

%% file: sec/2_method.tex
\section{System Design and Construction}
\label{sec:method}
Building an AI system with dynamic structure is challenging, especially when further trying to configure cost and performance. To tackle this, we emphasize a modular design in the proposed system. All experts are independently developed, and gating function could be implemented as separate add-on modules. This modular approach offers advantages over monolithic models, such as easier of leveraging existing resources, flexible deployment, and the potential for configurable efficiency and speculative execution. Although the system architecture is simple, its implementation requires task-specific designs. This section outlines the key principles for designing each module under different conditions.
\subsection{Runtime Architecture}
As shown in \cref{fig:overview}, the proposed system consists of $N_{\mathit{exp}}$ expert nodes, where $N_{\mathit{exp}} > 1$. Each node, except the last, contains an expert model with a post expert gate. All experts are frozen models pre-trained on the same task, and the gates generate a confidence value $c$ indicating whether the expert can provide an acceptable output for a given input. After computing the $i^{th}$ node, the confidence is compared with a pre-defined threshold $t_2^{(i)}$. If $c>t_2^{(i)}$, the entire computation terminates and the final output is the output of the $i^{th}$ expert. Otherwise, the computation continues with the next node. However, a visible issue with this architecture is the difficult samples need to waste computation on many small nodes which are not able to process them. We introduce the pre-expert gating mechanism to tackle this issue. When a sample reaches an intermediate $\mathit{Node}^{(j)}$, the confidence generated by the last computed node is compared with a pre-expert threshold $t_1^{(j)}$. If the confidence from the previous node is lower than $t_1^{(j)}$, it implies that this sample is too difficult for $\mathit{Node}^{(j)}$. The $\mathit{Node}^{(j)}$ is directly skipped in this case. This procedure is designed to terminate the computation with a confident output as early as possible.

\subsection{Experts}
The performance of the proposed system relies on the coordinated use of pre-trained models. Consequently, we assume that the employed experts are well-trained and have generalized effectively to their respective tasks. Their efficiency is also expected to be optimal under specific conditions. Fortunately, the rapid growth of the AI industry has led to the availability of numerous model candidates. Although current AI models are not perfect, most widely used models generally meet these expectations. ORXE has been designed to accommodate the imperfections of the experts; however, we still expect that all experts perform competently in their specific tasks.\\
The system requires experts to be ordered by model sizes, with each subsequent expert offering improved performance at a higher computational cost. This can guarantee that unconfident predictions are always handled a more powerful expert. Arranging state-of-the-art models in order without much selection work meets these requirements. The configuration search method, discussed in \cref{sec:config_generation}, automatically determines which experts of them should be activated.

\subsection{Gates}
Gates or routers are widely used in many deep learning designs \cite{cai_survey_2024}. They evaluate which branch is most beneficial for making accurate predictions. Typically, they are implemented as internal components of the model and trained jointly with the model, resulting in gating decisions that are often opaque and not externally configurable. In the proposed work, we design gates to generate confidence scores that quantify the reliability of an expert's prediction for a given input. The gating decision is then made by comparing this score against a defined threshold. Higher thresholds lead to a greater likelihood of using larger, more powerful experts, whereas lower thresholds allow the inference process to terminate early, reducing computational costs. This explicit thresholding approach makes the gating process both explainable, by providing interpretable confidence scores, and configurable, through the adjustment of thresholds. This section details the principles of gate design, and the method for determining thresholds is described in \cref{sec:config_generation}.
\paragraph{Confidence Score}
One important factor of an early exiting system is to produce reliable confidence scores to represent the prediction quality of inputs. A higher score indicates the prediction is more likely to be accepted. This indicator should ideally correlate to the evaluation metric. For example, in classification tasks, the predicted probability (i.e. the largest output value after the Softmax transformation) can be used as the confidence score. Meanwhile, the gating mechanism in early exiting is fundamentally designed to select the subset of samples having higher relative confidence. It primarily focuses on the relative differences in confidence scores across samples rather than their absolute calibration with actual performance metrics. Consequently, even if the model's predicted confidences are not well-calibrated, they can still effectively serve as reliable quality indicators within an early exiting framework. This finding supports employing the ORXE metamodel in classification tasks without additional calibration training for each expert. Utilizing native confidence not only eliminates the requirement for training but also reduces computational overhead associated with confidence estimation during inference. Other studies\cite{wang_tiny_2025} have similarly demonstrated the effectiveness and efficiency of leveraging native confidence for early exiting decisions in classification tasks.\\
However, for other tasks such as object detection, segmentation, or regression, models typically do not directly produce native confidence estimates for individual input samples. In such scenarios, it becomes essential to train an explicit confidence evaluator. We discuss potential calibration strategies for models lacking native confidence scores in \cref{sec:trained_conf}.
\section{Generating Configurations}
\label{sec:config_generation}
The configuration determines the overall behavior of the metamodel. Assigning a low threshold to an early node encourages more samples to exit with a smaller expert, thus significantly reducing the average inference cost at the expense of overall performance. Conversely, increasing the thresholds results in higher computational costs but improves overall performance. The configurations can be viewed as the learnable parameters of the metamodel. Due to the thresholding operation is not differentiable, these parameters cannot be learned by backward propagation methods. Therefore, in this section, we propose a search-based method to learn those parameters on a given dataset. Similar to other learning algorithms, this method also considers the computing efficiency, convergence and the overfitting issue. This algorithm can generate sets of threshold values to fulfil different requirements in practice.\\
To determine the optimal configuration for different scenarios, we first define the configuration of the ORXE system as a tuple:
\begin{align}
    \mathbf{g_{\lambda}}=(\mathbf{t}_{1,\lambda}, \mathbf{t}_{2,\lambda})
\end{align}
Where $\mathbf{t}_{1,\lambda} \in [0,1]^{N_{\mathit{exp}}-1}$ and $\mathbf{t}_{2,\lambda} \in [0,1]^{N_{\mathit{exp}}-1}$ are threshold vectors, and $t_1^{(i)}$ and $t_2^{(i)}$ are the thresholds for pre and post expert gating, respectively. The parameter $\lambda \in [0,1]$ represents the trade-off preference, where $\lambda=0$ corresponds to minimal cost and performance, while $\lambda=1$ corresponds to maximal cost and performance. The collection of configurations, $G=\{\mathbf{g}_{0.00}, \mathbf{g}_{0.01}, \cdots, \mathbf{g}_{\lambda}, \cdots, \mathbf{g}_{1.00}\}$, includes configurations that reflect different cost and performance preferences. This is the target of the configuration search process.\\
Given $\mathbf{t}_1 \in [0,1]^{N_{\mathit{exp}}-1}$ and $\mathbf{t}_2 \in [0,1]^{N_{\mathit{exp}}-1}$, along with the search step size $\delta$, the size of the entire search space is $N_g \left( \frac{1}{\delta} \right)^{2(N_{\mathit{exp}}-1)}$. This can be an enormous number, particularly when $N_g$ and $N_{\mathit{exp}}$ are large. DyCE\cite{wang_dyce_2024} introduced a heuristic search algorithm to generate a series of optimal configurations for a given dataset more efficiently. However, the problem of overfitting configurations persists. The effectiveness of cost-saving depends heavily on the dataset, and the evaluation gate may be less reliable in the test environment, especially for out-of-distribution samples. Consequently, the optimal configuration for one dataset may not be optimal for another. Therefore, the best practice involves collecting data from the production environment and updating the configuration accordingly, but high-quality dataset might be expensive to obtain.\\
Given these challenges, we assume that only the training set is available for configuration generation, and we introduce regularization into the search process to mitigate overfitting. Specifically, the generation process is divided into three steps: 1) First, we set $\mathbf{t}_1 = \mathbf{0}^{N_{\mathit{exp}}-1}$ and search for a limited number of optimal $\mathbf{t}_2$ on the training set, incorporating a regularization factor during this step. 2) We then remove suboptimal configurations and interpolate between every two adjacent remaining configurations. This approach reduces the search frequency and mitigates the risk of overfitting. 3) Thirdly, we search for the optimal $\mathbf{t}_{1,\lambda}$ for each generated $\mathbf{t}_{2,\lambda}$. The pre-expert gate does not directly impact cost and performance. Any values of $t_2^{(i)}>0$ will move samples to more powerful experts, generally without negatively affecting system performance. Therefore, searching for $\mathbf{t}_1$ should not lead to significant overfitting. 4) Finally, we remove any configurations which violate the monotonicity between the system performance and $\lambda$. This step can improve the generalization while tuning performance in unknown environments.

\subsection{Searching Thresholds for Post Expert Gating}
\begin{footnotesize}
    \begin{align}
        f(\mathbf{g}_\lambda)           & = (1-\lambda)\cdot \mathit{Cost}(\mathbf{g}_\lambda)+\lambda \cdot (1-\mathit{Perf}(\mathbf{g}_\lambda)) \label{eq:opt_obj}                           \\
     \mathit{Cost}(\mathbf{g}_\lambda) & = \frac{1}{N_{data}}  \sum_{i  = 0}^{N_{\mathit{exp}}} N_{comp}^{(i)}(\mathbf{g}_\lambda)\mathit{Cost}^{(i)}\\
\mathit{Perf}(\mathbf{g}_\lambda) & = \frac{1}{N_{data}}\sum_{i  = 0}^{N_{\mathit{exp}}} N_{exit}^{(i)}(\mathbf{g}_\lambda)\mathit{Perf}^{(i)}(\mathbf{g}_\lambda)reg(\lambda) \\
reg(\lambda)&=(1-(\alpha\lambda+\beta)(1-\overline{\mathit{Perf}}^{(i)}))\label{eq:perf_decay}
    \end{align}
\end{footnotesize}
The search methodology employed in this work is adapted from DyCE \cite{wang_dyce_2024}. We aim to determine the optimal threshold vector $\mathbf{t}_2$ by minimizing the objective function presented in \cref{eq:opt_obj}. This objective function is defined as a sum of the system's cost and performance, weighted by the preference factor $\lambda$. Specifically, $\mathit{Cost}(\mathbf{g}_\lambda)$ and $\mathit{Perf}(\mathbf{g}_\lambda)$ represent the normalized cost and performance metrics of the system under configuration $\mathbf{g}_\lambda$. The term $N_{comp}^{(i)}(\mathbf{g}_\lambda)$ indicates the number of samples processed by the $\mathit{Expert}^{(i)}$, while $N_{exit}^{(i)}(\mathbf{g}_\lambda)$ represents the number of samples exited at $\mathit{Node}^{(i)}$.

Optimizing \cref{eq:opt_obj} involves finding the optimal $\mathbf{t}_2$ on the training set. However, smaller experts tend to be less reliable when high performance is desired, often exhibiting overconfidence. This false high confidence can cause the search process to select these smaller experts by mistake. To address this, we introduce a regularization term, as defined in \cref{eq:perf_decay}, into the search process. The penalty strength depends on the expert's overall performance, the preference factor $\lambda$, and two coefficients, $\alpha$ and $\beta$. Here, $\mathit{Perf}^{(i)}$ represents the performance of $\mathit{Expert}^{(i)}$ on a subset of samples that were exited by $\mathit{Expert}^{(i)}$, while $\overline{\mathit{Perf}}^{(i)}$ is the statistical performance of $\mathit{Expert}^{(i)}$ on the entire training set. This approach ensures that smaller experts are penalized more heavily when high performance is prioritized, thereby mitigating overfitting caused by overconfident experts.\\
The overall inference cost is computed as the weighted sum of the computational costs incurred by each sample. $\mathit{Cost}^{(i)}$ represents the cost required to compute $\mathit{Node}^{(i)}$. This cost metric can be expressed as FLOPs for theoretical analysis but should be replaced with practical metrics such as latency or energy consumption in real-world systems. Given the heterogeneity of independent models, specific devices may prefer using a particular collection of experts to construct the ORXE system. \\
The minimization of \cref{eq:opt_obj} requires iterative experimentation with different threshold values. Notably, the objective function is typically convex with respect to each $t^{(i)}_{2}$, making the optimization process relatively straightforward. Further details regarding the search process are provided in \cref{sec:proof_of_convergence} of the appendix.

\subsection{Config Post-Processing}

In the previous step, we fixed all elements in $\mathbf{t}_1$ to zero and derived the optimal $\mathbf{t}_2$. However, due to discrepancies between the training set and the test set, the derived $\mathbf{t}_2$ may not necessarily be optimal in practice. Specifically, there could be an alternative $\mathbf{t}_{2,\lambda}^{\prime}$ such that $\mathit{Cost}(\mathbf{t}_{2,\lambda}^{\prime}) < \mathit{Cost}(\mathbf{t}_{2,\lambda})$ and $\mathit{Perf}(\mathbf{t}_{2,\lambda}^{\prime}) > \mathit{Perf}(\mathbf{t}_{2,\lambda})$. For every $\mathbf{t}_{2,\lambda}$ found on the training set, we discard it if such a $\mathbf{t}_{2,\lambda}^{\prime}$ exists, as these are typically erroneous results produced by the search process. After filtering out suboptimal configurations, we interpolate between each pair of adjacent remaining configurations. This interpolation reduces the frequency of the search process and thereby mitigates the risk of overfitting. Given two adjacent $\mathbf{t}_{1,\lambda}$, $\mathbf{t}_{1,\lambda+s}$ and a new preference $\lambda<\lambda^{\prime}<\lambda+s$, we can interpolate $\mathbf{t}_{1,\lambda^{\prime}}$ as follows:
\begin{align}
    \mathbf{t}_{1,\lambda ^ \prime} & = \mathbf{t}_{1,\lambda}+\frac{\lambda ^ \prime-\lambda}{s}( \mathbf{t}_{1,\lambda+s}- \mathbf{t}_{1,\lambda})
\end{align}
\subsection{Setting Pre-Expert Gating}
The purpose of the pre expert gating is to avoid unnecessary computation on samples that are evidently too challenging. In the proposed system, the confidence used for pre-expert gating is derived from the most recently computed node. This design relies on the assumption that if a sample is extremely difficult for a certain expert, it is likely to remain relatively difficult for larger expert models as well. Such a mechanism allows the most challenging samples to bypass intermediate nodes, thereby reducing inference latency in worst cases.\\
The search process for $\mathbf{t}_{1,\lambda}$ follows a similar method to that used for $\mathbf{t}_{2,\lambda}$, except that $\mathbf{t}_{2,\lambda}$ is fixed to the previously generated value while $\mathbf{t}_{1,\lambda}$ is optimized by the searching algorithm.
\subsection{Monotonicity}
\label{sec:mono}
Note that ORXE does not strictly guarantee monotonicity of model performance or computational cost w.r.t the preference parameter $\lambda$. Within a limited range, increasing $\lambda$ may occasionally lead to a decrease in both performance and cost. This phenomenon is similar to situations in which a model outperforms another on one dataset but does not necessarily perform better on other datasets, especially when the two models are close in capacities. Nevertheless, when $\lambda$ is increased substantially, ORXE consistently results in significant performance improvements.\\
To enhance the correlation between $\lambda$ and overall system performance on unseen datasets, configurations generated through the calibration dataset should undergo an additional filtering step. Specifically, configurations are first sorted in ascending order based on their associated $\lambda$ values. Iterating from the smallest to largest $\lambda$, we remove any configuration $\mathbf{g}_{\lambda_2}$ for which there exists a smaller $\lambda_1 < \lambda_2$ such that the following condition is violated:
{\small
\begin{align}
\mathit{Conf}(\mathbf{g}_{\lambda_1}) < \mathit{Conf}(\mathbf{g}_{\lambda_2}) \land \mathit{Cost}(\mathbf{g}_{\lambda_1}) < \mathit{Cost}(\mathbf{g}_{\lambda_2})
\end{align}
}
This address monotonicity in terms of both confidence and computational cost across configurations, thereby improving generalization performance and reliability.

%% file: sec/3_expriment.tex
\section{Experiments}
\label{sec:experiment}
\subsection{Model Construction}
The construction of the ORXE metamodel requires a clear definition of cost and performance metrics, based on the specific task and application. An efficient engine is also needed to track system cost and performance as configurations change to accelerate the search process. Additionally, ORXE relies on the availability of well-trained models for the given task. Thus, classification tasks being most suitable due to their extensive study and numerous pre-trained models. The native probability output from classification models provides a strong indicator of prediction quality, allowing the formation of the ORXE system \textbf{without any additional training}.\\
In this paper, we implemented ORXE across multiple devices for classification tasks. We gathered over 60 state-of-the-art models from the timm library\cite{rw2019timm}, recording their predictions for each image in the ImageNet-1k\cite{russakovsky_imagenet_2015} dataset. We also measured the FLOPs and inference latency of each candidate expert on different devices. The experts were shortlisted based on their efficiency on specific devices with ImageNet validation set. It ensures that each selected model demonstrated superior performance compared to models with similar computational costs. These shortlisted models were then arranged in ascending order of computational cost and validation accuracy. Since model costs vary across devices, the shortlist and ordering differed for each target device.\\
During the configuration generation, we tracked the pathway of each image through the nodes, including the final exit point. As the cost of each expert and the correctness of each image's classification were pre-determined, we could track overall cost and performance accordingly. We then followed the procedure described in \cref{sec:config_generation} to generate configurations using the training set. The metamodel was composed of 20 heterogeneous models, ranging from small to large, providing a broad range of cost and performance trade-offs. The results on the validation set are reported in subsequent sections, with additional implementation details provided in \cref{sec:implementation_details} of the appendix.

\begin{figure}[htbp]
  \centering
  \begin{subfigure}{\linewidth}
    \includegraphics[width=\linewidth]{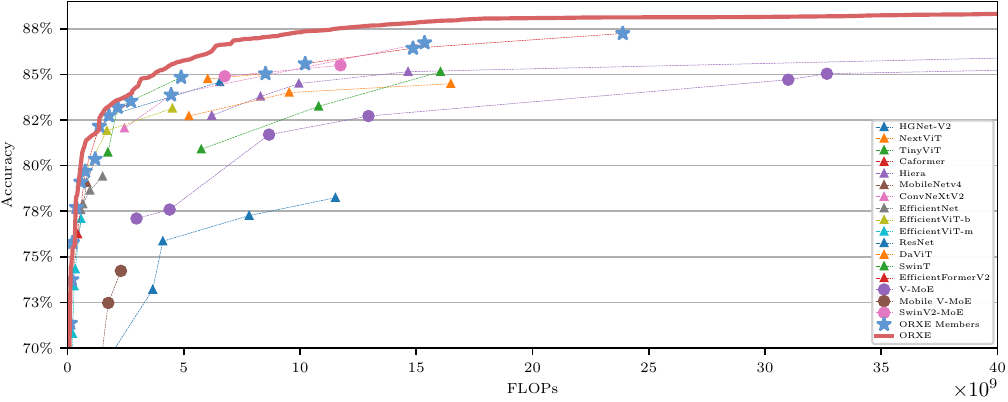}
    \caption{Comparison of ORXE metamodels with state-of-the-art models on ImageNet validation set. The model marked by blue stars are the members of the metamodel. }
    \label{fig:compare_with_static}
  \end{subfigure}
  \\
  \begin{subfigure}{\linewidth}
    \includegraphics[width=1\textwidth]{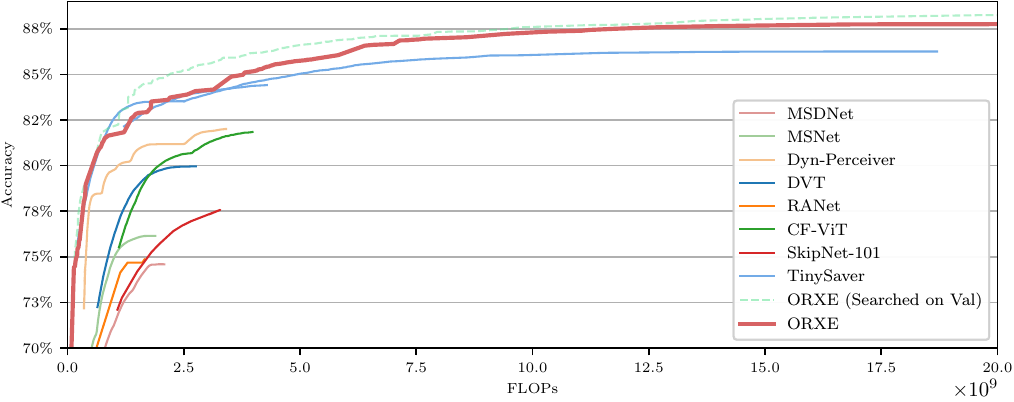}
	\caption{Comparison of ORXE metamodels with other models that enables configurable efficiency.}
	\label{fig:compare_with_dynamic}
  \end{subfigure}
  \caption{Comparison of ORXE metamodels with related work on ImageNet Val}

\end{figure}
\subsection{System Efficiency}
In the experiment, we initially use FLOPs as the cost metric to construct the ORXE system. Configurations are generated based on the ImageNet training set, and the average FLOPs per sample and corresponding accuracy on the Validation set are reported. As shown in \cref{tab:flops}, the ORXE metamodel generally achieves equivalent accuracy with significantly fewer FLOPs. Notably, we are not compared with vanilla models, but their versions boosted by the community with engineering tricks. In practice, most samples are relatively easy and can be accurately classified by smaller models. Even though the criterion for distinguishing sample difficulty, comparing native confidence to a threshold, is not perfect, the overall computational cost can still be reduced considerably using this approach. \cref{fig:compare_with_static} provides a more intuitive illustration of these results. ORXE demonstrates superior efficiency compared to many state-of-the-art models, including any individual member of the metamodel. Moreover, the constituent models within ORXE are replaceable. If more advanced models are developed in the future, they can be incorporated into ORXE to further enhance its efficiency.
\begin{table}[htbp]
\caption{Comparison of ORXE metamodels with state-of-the-art models on ImageNet validation set. The input size to all models is $224 \times 224$. The ORXE approach can have less FLOPs than standalone models with similar accuracy.}
\label{tab:flops}
\resizebox{\linewidth}{!}{%
\begin{tabular}{l|lll}
\hline
Model                                                        & Acc              & FLOPs (G)     & Params (M)       \\ \hline
\textbf{ORXE ($\lambda=0.995$)}                              & \textbf{88.35\%} & \textbf{45.3} & \textbf{516.3}   \\
BeiT$_\mathit{large}$\cite{peng_beit_2022}                   & 88.34\%          & 61.6          & 304.4            \\ \hline
\textbf{ORXE ($\lambda=0.793$)}                              & \textbf{87.34\%} & \textbf{10.3} & \textbf{815.6}   \\
CAFormer$_\mathit{b36}$\cite{yu_metaformer_2023}             & 87.25\%          & 23.9          & 211.9            \\ \hline
\textbf{ORXE ($\lambda=0.785$)}                              & \textbf{86.91\%} & \textbf{7.5}  & \textbf{815.6}   \\
Hiera$_\mathit{huge}$\cite{ryali_hiera_2023}                 & 86.90\%          & 127.9         & 949.0            \\
ConvNeXtV2$_\mathit{base}$\cite{woo_convnext_2023-1}         & 86.74\%          & 15.4          & 152.3            \\ \hline
\textbf{ORXE ($\lambda=0.620$)}                              & \textbf{85.17\%} & \textbf{4.0}  & \textbf{1,008.7} \\
Swin$_\mathit{base}$\cite{liu_swin_2021-1}                   & 85.14\%          & 16.0          & 187.7            \\
Next-ViT$_\mathit{base}$\cite{li_next-vit_2022}              & 85.05\%          & 8.5           & 170.6            \\
TinyViT$_\mathit{21m}$\cite{wu_tinyvit_2022}                 & 84.84\%          & 4.9           & 96.5             \\
MaxViT$_\mathit{large}$\cite{tu_maxvit_2022}                 & 84.83\%          & 45.6          & 352.1            \\
HGNet-V2$_\mathit{b5}$\cite{cui_beyond_2021}                 & 84.59\%          & 6.5           & 170.6            \\
DaViT$_\mathit{base}$\cite{ding_davit_2022}                  & 84.49\%          & 16.5          & 333.6            \\ \hline
XCiT$_\mathit{small}$\cite{el-nouby_xcit_2021}               & 83.98\%          & 21.1          & 96.5             \\
\textbf{ORXE ($\lambda=0.387$)}                              & \textbf{83.90\%} & \textbf{2.6}  & \textbf{639.1}   \\
EfficientFormerV2$_\mathit{l}$\cite{li_rethinking_2023}      & 83.53\%          & 2.7           & 147.0            \\
EfficientViT$_\mathit{b3}$\cite{cai_efficientvit_2023}       & 83.14\%          & 4.5           & 187.7            \\
\textbf{ORXE ($\lambda=0.199$)}                              & \textbf{81.65\%} & \textbf{1.0} & \textbf{1,385.9} \\
MobileNetV4$_\mathit{hybrid-m}$\cite{qin_mobilenetv4_2024}   & 80.36\%          & 1.2           & 342.1            \\ \hline
\textbf{ORXE ($\lambda=0.200$)}                              & \textbf{78.95\%} & \textbf{0.4}  & \textbf{257.7}   \\
ResNet$_\mathit{152}$\cite{he_deep_2016}                     & 78.24\%          & 11.5          & 464.9            \\
EfficientViT$_\mathit{m5}$\cite{liu_efficientvit_2023}       & 77.08\%          & 0.6           & 75.9             \\
\textbf{ORXE ($\lambda=0.055$)}                              & \textbf{75.78\%} & \textbf{0.2}  & \textbf{115.3}   \\
MobileNetV3$_\mathit{large_100}$\cite{howard_searching_2019} & 75.78\%          & 0.2           & 70.0             \\
\textbf{ORXE ($\lambda=0.020$)}                              & \textbf{74.42\%} & \textbf{0.1}  & \textbf{149.7}   \\
MobileNetV3$_\mathit{small_100}$\cite{howard_searching_2019} & 67.64\%          & 0.1           & 34.4             \\
EfficientViT$_\mathit{m0}$\cite{liu_efficientvit_2023}       & 63.27\%          & 0.1           & 75.9             \\ \hline
\end{tabular}%
}
\end{table}
The proposed method not only conserves computational resources but also provides significant flexibility in tuning the trade-off between cost and performance. As illustrated in \cref{fig:compare_with_static}, the implemented ORXE model demonstrates adjustable accuracy ranging from 70\% to over 88\%. This capability allows the system to adapt precisely to varying real-time requirements by switching between configurations. In practical scenarios, system performance can be slightly reduced when inference resources, such as battery level or available computational units, are limited. Once these resources are restored, the performance can be instantly adjusted back to higher levels. This adaptability ensures that the user experiences minimal disruption while maintaining consistent system availability.\\
The combination of multiple models in the ORXE framework also contributes to enhanced performance. As shown in \cref{tab:flops}, the ORXE system achieves higher accuracy compared to its largest individual expert. Specifically, the $\mathit{BeiTv2_{large}}$ model \cite{peng_beit_2022} achieves an accuracy of 88.34\% with 61.6G FLOPs, whereas the ORXE model reaches an accuracy of 88.35\% with only 45.3G FLOPs.\\
However, the sacrifice for this high flexibility and enhanced efficiency is the increased number of parameters required. The metamodel typically requires multiple models to work together, necessitating the loading of multiple models simultaneously. The complete ORXE classification model, which consists of 20 experts, has a total of 795M parameters. Most of these parameters are concentrated in the largest models. In practice, if the application requires a narrower range of performance, the metamodel can be tailored to meet those specific needs with fewer parameters. \cref{tab:flops} also displays the number of parameters required for different configurations. Importantly, since each node in ORXE operates independently, individual experts can be deployed across different devices. This modularity means that the additional parameters do not pose significant challenges, provided each expert can fit into the respective inference device.\\
\cref{fig:compare_with_dynamic} presents a comparison between ORXE and other dynamically configurable models, including early exiting and layer skipping models. The ORXE metamodel demonstrates a clear advantage in most scenarios, particularly in achieving high accuracy. Unlike other approaches that typically require intricate designs and specialized training procedures to incorporate dynamic efficiency features, ORXE avoids these complexities. Such design burdens often hinder other models from focusing on performance and scaling effectively to high-performance levels. In contrast, the control logic and expert units are decoupled in ORXE. This decoupling allows performance and flexibility to be optimized independently without interfering with each other.\\
We also provide the ORXE performance curve obtained when configurations are searched on the validation set, offering a comparison to illustrate a theoretical upper bound of the proposed metamodel. This performance is approachable when a high-quality dataset is available for config generation.
\subsection{On-device Test}
\label{sec:on_device_test}
\begin{figure}[bp]
  \centering
  \begin{subfigure}{0.45\linewidth}
    \includegraphics[width=\linewidth]{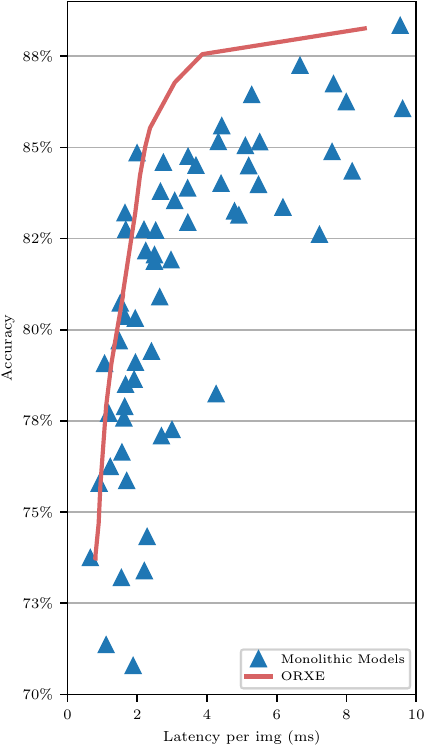}
    \caption{RTX 3090, BS1}
    \label{fig:speed_test:gpuonnx1}
  \end{subfigure}
  \begin{subfigure}{0.45\linewidth}
    \includegraphics[width=1\textwidth]{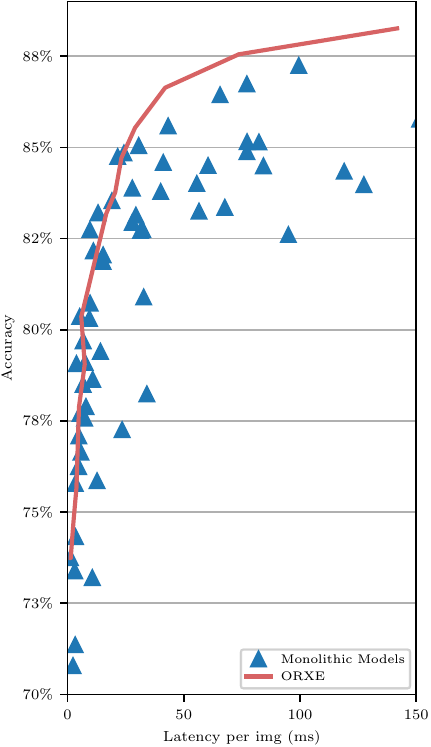}
	\caption{CPU, BS1}
	\label{fig:speed_test:cpu1}
  \end{subfigure}
    \caption{The speed test of ORXE and other models on different GPU and CPU with ONNX Runtime\cite{onnxruntime}}
    \label{fig:speed_test}
\end{figure}

The FLOPs analysis, while informative, cannot substitute for practical testing. We validate the ORXE methodology across multiple devices under varying deployment configurations. It is important to note that the efficiency of the same model can vary significantly between different devices. Consequently, we measure the practical inference latency per image for each model across different devices, using these latency values as the cost metric to construct the ORXE model. The ORXE model is then evaluated on the same devices, with the results presented subsequently. \cref{fig:speed_test} illustrates that, on CPU and GPU. The ORXE model outperforms other models, in most cases, while obtaining the characteristic of high dynamic range. The test on more deployment cases is provided in \cref{sec:additional_experiments}\\
\subsection{Generalization}
\begin{figure}[htbp]
  \centering
  \begin{subfigure}{\linewidth}
    \includegraphics[width=\linewidth]{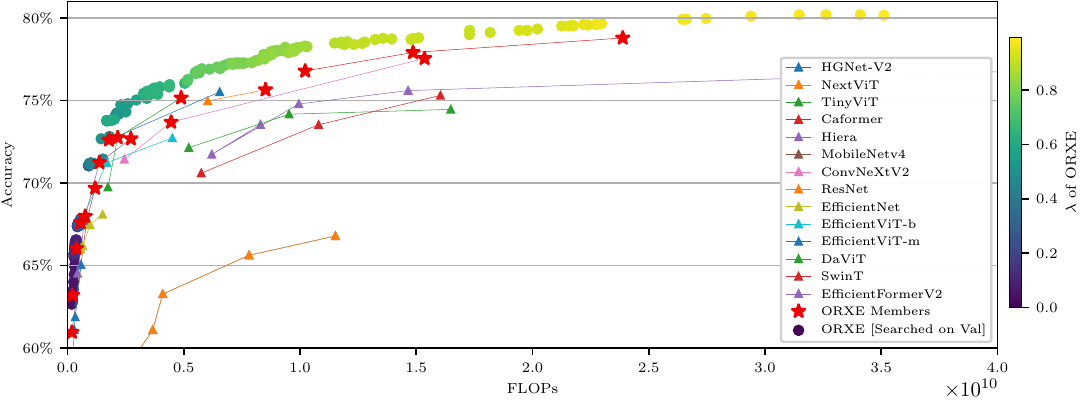}
    \caption{ImageNetv2\cite{recht_imagenet_2019}}
    \label{fig:eval:independent:imv2}
  \end{subfigure}
  \\
  \begin{subfigure}{\linewidth}
    \includegraphics[width=1\textwidth]{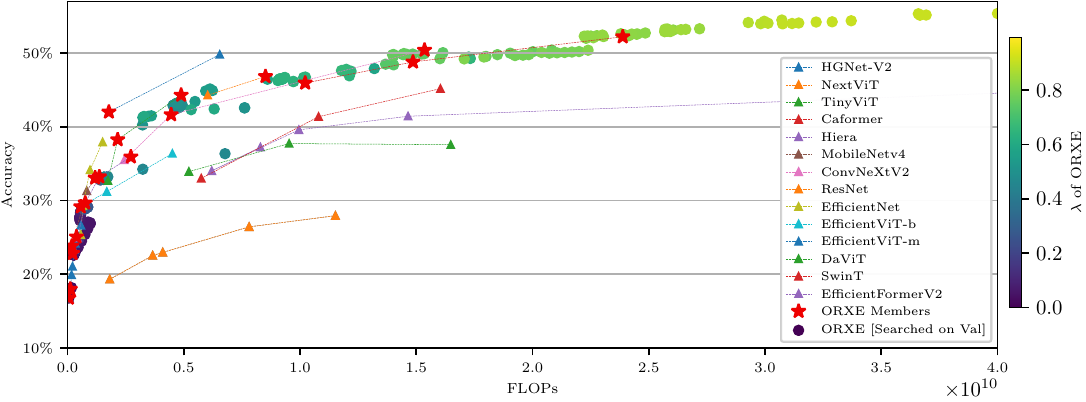}
	\caption{ImageNet-Sketch\cite{wang_learning_2019}}
	\label{fig:eval:independent:ims}
  \end{subfigure}

	\caption{Evaluation results on independent datasets}
	\label{fig:eval:independent}
\end{figure}

In earlier sections, we demonstrated the performance of ORXE on the ImageNet validation set. However, during development, hyperparameter tuning and expert selection leveraged information from the validation set itself. To properly evaluate the generalization capability of ORXE, we employed additional independent datasets. Specifically, ImageNetV2\cite{recht_imagenet_2019} is a dataset closely matching the distribution of ImageNet, containing 10,000 distinct images, whereas ImageNet-Sketch\cite{wang_learning_2019} is a domain shifted out-of-distribution (OOD) dataset consisting of 50,000 sketch-based images rather than natural photographs.\\
Configurations generated from ImageNet Val were filtered as described in \cref{sec:mono} then applied to these datasets, and the corresponding results are presented in \cref{fig:eval:independent}. The proposed method generalizes well on ImageNetV2, consistently surpassing the performance of individual expert models with similar FLOPs in most scenarios. Furthermore, a clear monotonic relationship between $\lambda$, performance, and cost is observed in the majority of cases. This finding indicates that $\lambda$ can effectively serve as a real-time performance tuning parameter during practical deployment.\\
In practice, the difficulty of incoming samples remains inherently unpredictable, preventing strict guarantees on performance or computational cost. However, a dynamic tuning mechanism can be introduced during deployment. Such a mechanism could periodically modify $\lambda$ in response to deviations of average computational costs from targeted constraints, thereby achieving controlled system inference cost.\\
As illustrated in \cref{fig:eval:independent:ims}, the performance of ORXE experiences noticeable degradation on OOD data, with several outliers appearing that weaken the correlation between $\lambda$ and performance. OOD data significantly reduces the effectiveness of the experts, as all experts were originally trained on natural images rather than sketches. Consequently, their confidence estimates become less reliable, violating a fundamental assumption of ORXE, which is each expert performs well on the target task. Additionally, the distribution shift between the calibration set and the test set limits the generalization of gating configurations. Despite these challenges, ORXE maintains its basic functionality, continues to scale performance across varying computational budgets, and does not significantly lag behind other models in terms of efficiency. In practical scenarios, naturally occurring OOD data appears infrequently, and therefore, it is unlikely to substantially impact the overall effectiveness of ORXE.
\section{Conclusion}
\label{sec:conclusion}
This paper introduces ORXE, a highly flexible framework designed to achieve real-time configurable efficiency across a wide range of scenarios without imposing significant additional overhead. Despite incorporating features of dynamic efficiency, the ORXE metamodel enables difficulty-aware compression using a simpler approach compared to existing methods, and demonstrates superior efficiency compared to most state-of-the-art models. The proposed system is both modular and adaptable, allowing for flexible construction and deployment across diverse applications. In this work, we discuss the design principles and construction strategies in different scenarios. We implemented the ORXE system for image classification tasks and validated its effectiveness through deployment on multiple devices. The underlying concept of ORXE can also be extended to other tasks with appropriate design adaptations.

%% file: sec/X_suppl.tex
\clearpage
\appendix
\setcounter{table}{0}
\renewcommand\thetable{\Alph{section}\arabic{table}}
\setcounter{figure}{0}
\renewcommand\thefigure{\Alph{section}\arabic{figure}}
\setcounter{page}{1}
\maketitlesupplementary

\section{Details of the Configuration Searching}
\label{sec:proof_of_convergence}
\subsection{Searching Process}
The configuration search is formulated as an optimization problem. The objective function \cref{eq:opt_obj} is not derivativable, but we can still optimize it. Multiple methods can be used. In this paper, we use the procedure described in \cref{alg:circular_search_algorithm}. This searching method starts with an empty config, i.e. $\mathbf{t}_{2,\lambda}=\mathbf{1}_{N_{exp}-1}$. It takes multiple rounds to modify the configuration. In each round, it will try to make modifications on every node, but only the best modification is adopted at the end of this round. Each modification attempt means minimizing the objective function w.r.t. a single element of $\mathbf{t}_{2,\lambda}$. \cref{eq:opt_obj} is convex w.r.t. every $\mathbf{t}_{2,\lambda}^{(i)}$, we prove this property in \cref{sec:convergence}. Therefore, this minimization can be achieved by any convex optimization method. After searching for the optimal $\mathbf{t}_{2,\lambda}$, we can fix it and search for the optimal $\mathbf{t}_{1,\lambda}$ in the same way. The convergence analysis of $\mathbf{t}_{1,\lambda}$ is more complicated but \cref{eq:opt_obj} is still empirically convex w.r.t. every $\mathbf{t}_{1,\lambda}^{(i)}$. Even if \cref{eq:opt_obj} is not strictly convex and the searching process may not converge to the global minimum, the searching process can work well in practice. Similar to the training process of a deep learning model, converge to the global minimum on the training set is not required and desired.
\begin{algorithm}[h]
	\KwIn{The preference factor, $\lambda \in [0,1]$.}
	\KwOut{Thresholds for all Gate 2, $\mathbf{t}_{2,\lambda}\in [0,1]^{N_{exp}-1}$.}
    Fix $\mathbf{t}_{1,\lambda}=\mathbf{0}_{N_{exp}-1}$, we search the threshold for gate 2 first.\\
	Initialize $\mathbf{t}_{2,\lambda}=\mathbf{1}_{N_{exp}-1}$, i.e. all samples go to the last by default.\\	
	Initialize object metric, $f_{min}= f_2(\mathbf{t}_{2,\lambda}) $.\\
	\While{$\mathrm{True}$} {
	    $\mathrm{Candidate} \leftarrow \mathrm{None}$\\
    	\For{$i=1 \rightarrow N$}	{
    		
            $\{t_c,~f_c\} \leftarrow \mathop{min}_{t^{(i)}_{2,\lambda}} f_2(\mathbf{t}_{2,\lambda})$ \\
            \If{$f_c < f_{min}$}{
                $\mathrm{Candidate} \leftarrow (i,~t_c)$\\
                $f_{min} \leftarrow f_c$
            }

    	}
    	\uIf{$\mathrm{Candidate} ~\text{is not} ~\mathrm{None}$}{
    	    $i,~t_c \leftarrow \mathrm{Candidate}$\\
			$t^{(i)}_{2,\lambda} \leftarrow t_c$ \\		
        }
        \Else{
        \textbf{break}
        }
        
    }
	\caption{Searching $\mathbf{t}_{2,\lambda}$}\label{alg:circular_search_algorithm}
\end{algorithm}
 
\subsection{Convergence}
\label{sec:convergence}
When $\mathbf{t}_{1,\lambda}$ is fixed as zero, the objective function is written as:
\begin{footnotesize}
    \begin{align}
        f(\mathbf{t}_{2,\lambda})           & = (1-\lambda)\cdot \mathit{Cost}(\mathbf{t}_{2,\lambda})+\lambda \cdot (1-\mathit{Perf}(\mathbf{t}_{2,\lambda}))                
    \end{align}
\end{footnotesize}

Where $\mathbf{t}_{2,\lambda} \in [0, 1]^N, \lambda \in [0, 1]$
Although $\mathit{Cost}(\mathbf{t}_{2,\lambda})$ and $\mathit{Perf}(\mathbf{t}_{2,\lambda})$ are not actually derivativable, we can still analyze their derivatives.
To prove that $f$ has a minimum, we firstly compute the partial derivative of $f$ with respect to $t^{(i)}_{2,\lambda}$ as
\begin{align}
    \frac{\partial f}{\partial t^{(i)}_{2,\lambda}} = (1-\lambda)\cdot \frac{\partial \mathit{Cost}}{\partial t^{(i)}_{2,\lambda}}-\lambda \cdot \frac{\partial \mathit{Perf}}{\partial t^{(i)}_{2,\lambda}}
    \label{eq:post_gate_f_derivative}
\end{align}
At an extremum, the partial derivative with respect to $t^{(i)}_{2,\lambda}$ must vanish.
\begin{align}
    \frac{\partial f}{\partial t^{(i)}_{2,\lambda}} & = (1-\lambda)\cdot \frac{\partial \mathit{Cost}}{\partial t^{(i)}_{2,\lambda}}-\lambda \cdot \frac{\partial \mathit{Perf}}{\partial t^{(i)}_{2,\lambda}} = 0       \\
                                        & \Rightarrow (1-\lambda)\cdot \frac{\partial \mathit{Cost}}{\partial t^{(i)}_{2,\lambda}}=\lambda \cdot \frac{\partial \mathit{Perf}}{\partial t^{(i)}_{2,\lambda}} \\
                                        & \Rightarrow   \frac{\partial \mathit{Perf}}{\partial \mathit{Cost}} =\frac{1-\lambda}{\lambda} \label{eq:post_gate_extrema}
\end{align}
Meanwhile, $\frac{\partial \mathit{Perf}}{\partial \mathit{Cost}}$ means slope of the cost - performance curve. In general, when we add more resource to a system, the performance will increase. However, due to the marginal effect, the increase of performance per cost unit will decrease. i.e.
\begin{align}
    \frac{\partial^2 \mathit{Perf}}{\partial \mathit{Cost}^2}<0 \label{eq:post_gate_2nd_derivative_lt_zero}
\end{align}
This phenomenon is also verified by some research on the scaling law of deep learning models. Additionally, $\frac{\partial \mathit{Perf}}{\partial \mathit{Cost}}$ is $\infty$ when $t^{(i)}_{2,\lambda}=0$ and approaching 0 when $t^{(i)}_{2,\lambda} \to \infty$. Therefore, \cref{eq:post_gate_extrema} has only one solution $t^{(i)}_{2,\lambda}>t^{(i)}_{2,\lambda,opt}$, which means $f$ has and only has one global extremum. Moreover, in our proposed system, we assume that every expert is placed in order. Therefore, the later nodes always have higher performance and more cost than the previous ones. For every $t^{(i)} \in \mathbf{t}_{2,\lambda}$, the increase of $t^{(i)}_{2,\lambda}$ will always make more samples processed by later experts. It means more overall cost and better performance, hence:
\begin{align}
    \frac{\partial \mathit{Cost}}{\partial t^{(i)}_{2,\lambda}}>0,\frac{\partial \mathit{Perf}}{\partial t^{(i)}_{2,\lambda}}>0\label{eq:post_gate_derivative_gt_zero}
\end{align}
Combining \cref{eq:post_gate_f_derivative,eq:post_gate_2nd_derivative_lt_zero,eq:post_gate_derivative_gt_zero}, we have:
\begin{align}
    \frac{\partial \mathit{Perf}}{\partial \mathit{Cost}}
    \begin{cases}
        < \frac{1-\lambda}{\lambda} \Rightarrow \frac{\partial \mathit{f}}{\partial \mathit{t^{(i)}_{2,\lambda}}} < 0 , & t^{(i)}_{2,\lambda}<t^{(i)}_{2,\lambda,opt} \\
        = \frac{1-\lambda}{\lambda}\Rightarrow \frac{\partial \mathit{f}}{\partial \mathit{t^{(i)}_{2,\lambda}}} = 0 ,  & t^{(i)}_{2,\lambda}=t^{(i)}_{2,\lambda,opt} \\
        >\frac{1-\lambda}{\lambda}\Rightarrow \frac{\partial \mathit{f}}{\partial \mathit{t^{(i)}_{2,\lambda}}} > 0 ,   & t^{(i)}_{2,\lambda}>t^{(i)}_{2,\lambda,opt}
    \end{cases}
\end{align}
$f(t^{(i)}_{2,\lambda})$ is decreasing first and then increasing at around $t^{(i)}_{2,\lambda,opt}$. Therefore, $f$ is convex at every dimension of $\mathbf{t}_{2,\lambda}$, which means that the minimization of $f$ with respect to $\mathbf{t}_{2,\lambda}$ can converge to a global minimum.

\section{Implementation Details}
\label{sec:implementation_details}

The ORXE metamodel consists of 20 models ranging from small to large, selected from about 70 well-trained models from timm\cite{rw2019timm}. For each selected model, we recorded predictions and confidence scores on the training set. Using the aforementioned method, we generated configurations with regularization coefficients $\alpha=2.0,~\beta=0.2$. The search step was set to 0.01, resulting in 100 configurations, which were then interpolated with a step of 0.001. Configurations that were clearly suboptimal were removed. Hundreds of configurations are available through this process. \cref{tab:flops_full} presents the complete version of \cref{tab:flops}, comparing ORXE with state-of-the-art models. The ORXE metamodel consistently demonstrates superior efficiency.

\begin{table*}[]
\caption{The full table of the FLOPs analysis of ORXE with the comparison of the state-of-the-art models. Models with underscore are employed as members of ORXE.}
\label{tab:flops_full}
\centering
\resizebox{!}{11cm}{%
\begin{tabular}{l|lll|l}
\hline
Model                                                              & Acc              & FLOPs (G)     & Params (M)       & timm\cite{rw2019timm} Checkpoint Name /   ORXE Preference \\
\textbf{ORXE}                                                      & \textbf{88.35\%} & \textbf{45.3} & \textbf{516.3}   & \textbf{$\lambda=0.995$}                                  \\
\textbf{ORXE}                                                      & \textbf{88.35\%} & \textbf{42.5} & \textbf{516.3}   & \textbf{$\lambda=0.994$}                                  \\
{\ul BeiT$_\mathit{large}$\cite{peng_beit_2022}}                   & 88.34\%          & 61.6          & 304.4            & \texttt{beitv2\_large\_patch16\_224.in1k\_ft\_in22k\_in1k}      \\ \hline
\textbf{ORXE}                                                      & \textbf{87.34\%} & \textbf{10.3} & \textbf{815.6}   & \textbf{$\lambda=0.793$}                                  \\
{\ul CAFormer$_\mathit{b36}$\cite{yu_metaformer_2023}}             & 87.25\%          & 23.9          & 211.9            & \texttt{caformer\_b36.sail\_in22k\_ft\_in1k}                  \\ \hline
\textbf{ORXE}                                                      & \textbf{86.91\%} & \textbf{7.5}  & \textbf{815.6}   & \textbf{$\lambda=0.785$}                                  \\
Hiera$_\mathit{huge}$\cite{ryali_hiera_2023}                       & 86.90\%          & 127.9         & 949.0            & \texttt{hiera\_huge\_224.mae\_in1k\_ft\_in1k}                  \\
{\ul ConvNeXtV2$_\mathit{base}$\cite{woo_convnext_2023-1}}         & 86.74\%          & 15.4          & 152.3            & \texttt{convnextv2\_base.fcmae\_ft\_in22k\_in1k}              \\
{\ul CAFormer$_\mathit{m36}$\cite{yu_metaformer_2023}}             & 86.44\%          & 14.9          & 211.9            & \texttt{caformer\_m36.sail\_in22k\_ft\_in1k}                  \\
Swin$_\mathit{large}$\cite{liu_swin_2021-1}                        & 86.24\%          & 34.9          & 265.0            & \texttt{swin\_large\_patch4\_window7\_224.ms\_in22k\_ft\_in1k}   \\
Hiera$_\mathit{large}$\cite{ryali_hiera_2023}                      & 86.06\%          & 45.6          & 949.0            & \texttt{hiera\_large\_224.mae\_in1k\_ft\_in1k}                 \\ \hline
\textbf{ORXE}                                                      & \textbf{85.78\%} & \textbf{5.1}  & \textbf{1,064.5} & \textbf{$\lambda=0.739$}                                  \\
ConvNeXtV2$_\mathit{large}$\cite{woo_convnext_2023-1}              & 85.77\%          & 34.4          & 342.1            & \texttt{convnextv2\_large.fcmae\_ft\_in1k}                   \\
{\ul CAFormer$_\mathit{s36}$\cite{yu_metaformer_2023}}             & 85.59\%          & 10.2          & 211.9            & \texttt{caformer\_s36.sail\_in22k\_ft\_in1k}                  \\
Hiera$_\mathit{baseplus}$\cite{ryali_hiera_2023}                   & 85.15\%          & 14.6          & 168.6            & \texttt{hiera\_base\_plus\_224.mae\_in1k\_ft\_in1k}             \\
Swin$_\mathit{base}$\cite{liu_swin_2021-1}                         & 85.14\%          & 16.0          & 187.7            & \texttt{swin\_base\_patch4\_window7\_224.ms\_in22k\_ft\_in1k}    \\
{\ul Next-ViT$_\mathit{base}$\cite{li_next-vit_2022}}              & 85.05\%          & 8.5           & 170.6            & \texttt{nextvit\_base.bd\_ssld\_6m\_in1k}                     \\ \hline
\textbf{ORXE}                                                      & \textbf{84.88\%} & \textbf{3.5}  & \textbf{704.2}   & \textbf{$\lambda=0.552$}                                  \\
{\ul TinyViT$_\mathit{21m}$\cite{wu_tinyvit_2022}}                 & 84.84\%          & 4.9           & 96.5             & \texttt{tiny\_vit\_21m\_224.dist\_in22k\_ft\_in1k}              \\
MaxViT$_\mathit{large}$\cite{tu_maxvit_2022}                       & 84.83\%          & 45.6          & 352.1            & \texttt{maxvit\_large\_tf\_224.in1k}                         \\
MaxViT$_\mathit{base}$\cite{tu_maxvit_2022}                        & 84.80\%          & 24.9          & 333.6            & \texttt{maxvit\_base\_tf\_224.in1k}                          \\
Next-ViT$_\mathit{small}$\cite{li_next-vit_2022}                   & 84.75\%          & 6.0           & 170.6            & \texttt{nextvit\_small.bd\_ssld\_6m\_in1k}                    \\
HGNet-V2$_\mathit{b5}$\cite{cui_beyond_2021}                       & 84.59\%          & 6.5           & 170.6            & \texttt{hgnetv2\_b5.ssld\_stage2\_ft\_in1k}                   \\
Hiera$_\mathit{base}$\cite{ryali_hiera_2023}                       & 84.50\%          & 9.9           & 168.6            & \texttt{hiera\_base\_224.mae\_in1k\_ft\_in1k}                  \\
DaViT$_\mathit{base}$\cite{ding_davit_2022}                        & 84.49\%          & 16.5          & 333.6            & \texttt{davit\_base}                                       \\
MaxViT$_\mathit{small}$\cite{tu_maxvit_2022}                       & 84.34\%          & 12.9          & 128.9            & \texttt{maxvit\_small\_tf\_224.in1k}                         \\
DaViT$_\mathit{small}$\cite{ding_davit_2022}                       & 84.01\%          & 9.5           & 464.9            & \texttt{davit\_small}                                      \\ \hline
XCiT$_\mathit{small}$\cite{el-nouby_xcit_2021}                     & 83.98\%          & 21.1          & 96.5             & \texttt{xcit\_small\_12\_p8\_224.fb\_dist\_in1k}                \\
\textbf{ORXE}                                                      & \textbf{83.90\%} & \textbf{2.6}  & \textbf{639.1}   & \textbf{$\lambda=0.387$}                                  \\
{\ul ConvNeXtV2$_\mathit{tiny}$\cite{woo_convnext_2023-1}}         & 83.88\%          & 4.5           & 152.3            & \texttt{convnextv2\_tiny.fcmae\_ft\_in22k\_in1k}              \\
Hiera$_\mathit{small}$\cite{ryali_hiera_2023}                      & 83.79\%          & 8.3           & 949.0            & \texttt{hiera\_small\_224.mae\_in1k\_ft\_in1k}                 \\
{\ul EfficientFormerV2$_\mathit{l}$\cite{li_rethinking_2023}}      & 83.53\%          & 2.7           & 147.0            & \texttt{efficientformerv2\_l.snap\_dist\_in1k}               \\
MaxViT$_\mathit{tiny}$\cite{tu_maxvit_2022}                        & 83.35\%          & 5.7           & 150.6            & \texttt{maxvit\_tiny\_tf\_224.in1k}                          \\
Swin$_\mathit{small}$\cite{liu_swin_2021-1}                        & 83.25\%          & 10.8          & 352.1            & \texttt{swin\_small\_patch4\_window7\_224.ms\_in22k\_ft\_in1k}   \\
{\ul TinyViT$_\mathit{11m}$\cite{wu_tinyvit_2022}}                 & 83.20\%          & 2.2           & 96.5             & \texttt{tiny\_vit\_11m\_224.dist\_in22k\_ft\_in1k}              \\
EfficientViT$_\mathit{b3}$\cite{cai_efficientvit_2023}             & 83.14\%          & 4.5           & 187.7            & \texttt{efficientvit\_b3.r224\_in1k}                        \\ \hline
\textbf{ORXE}                                                      & \textbf{82.93\%} & \textbf{1.7}  & \textbf{793.5}   & \textbf{$\lambda=0.289$}                                  \\
Hiera$_\mathit{tiny}$\cite{ryali_hiera_2023}                       & 82.74\%          & 6.2           & 168.6            & \texttt{hiera\_tiny\_224.mae\_in1k\_ft\_in1k}                  \\
{\ul HGNet-V2$_\mathit{b3}$\cite{cui_beyond_2021}}                 & 82.74\%          & 1.8           & 170.6            & \texttt{hgnetv2\_b3.ssld\_stage2\_ft\_in1k}                   \\
\textbf{ORXE}                                                      & \textbf{82.74\%} & \textbf{1.4}  & \textbf{636.1}   & \textbf{$\lambda=0.242$}                                  \\
DaViT$_\mathit{tiny}$\cite{ding_davit_2022}                        & 82.72\%          & 5.2           & 464.9            & \texttt{davit\_tiny}                                       \\
XCiT$_\mathit{tiny}$\cite{el-nouby_xcit_2021}                      & 82.61\%          & 12.3          & 96.5             & \texttt{xcit\_tiny\_24\_p8\_224.fb\_dist\_in1k}                 \\
{\ul EfficientFormerV2$_\mathit{s2}$\cite{li_rethinking_2023}}     & 82.16\%          & 1.4           & 187.7            & \texttt{efficientformerv2\_s2.snap\_dist\_in1k}              \\
ConvNeXtV2$_\mathit{nano}$\cite{woo_convnext_2023-1}               & 82.05\%          & 2.4           & 152.3            & \texttt{convnextv2\_nano.fcmae\_ft\_in22k\_in1k}              \\ \hline
EfficientViT$_\mathit{b2}$\cite{cai_efficientvit_2023}             & 81.91\%          & 1.7           & 265.0            & \texttt{efficientvit\_b2.r224\_in1k}                        \\
\textbf{ORXE}                                                      & \textbf{81.83\%} & \textbf{1.2}  & \textbf{1,092.0} & \textbf{$\lambda=0.192$}                                  \\
Swin$_\mathit{tiny}$\cite{liu_swin_2021-1}                         & 80.90\%          & 5.8           & 333.6            & \texttt{swin\_tiny\_patch4\_window7\_224.ms\_in22k\_ft\_in1k}    \\
\textbf{ORXE}                                                      & \textbf{80.76\%} & \textbf{0.6}  & \textbf{787.5}   & \textbf{$\lambda=0.142$}                                  \\
TinyViT$_\mathit{5m}$\cite{wu_tinyvit_2022}                        & 80.73\%          & 1.7           & 96.5             & \texttt{tiny\_vit\_5m\_224.dist\_in22k\_ft\_in1k}               \\
{\ul MobileNetV4$_\mathit{hybrid-m}$\cite{qin_mobilenetv4_2024}}   & 80.36\%          & 1.2           & 342.1            & \texttt{mobilenetv4\_hybrid\_medium.e500\_r224\_in1k}         \\ \hline
{\ul EfficientFormerV2$_\mathit{s1}$\cite{li_rethinking_2023}}     & 79.69\%          & 0.8           & 147.0            & \texttt{efficientformerv2\_s1.snap\_dist\_in1k}              \\
EfficientNet$_\mathit{b4}$\cite{tan_efficientnet_2019-1}           & 79.41\%          & 1.5           & 128.9            & \texttt{efficientnet\_b4.ra2\_in1k}                         \\
{\ul EfficientViT$_\mathit{b1}$\cite{cai_efficientvit_2023}}       & 79.10\%          & 0.6           & 187.7            & \texttt{efficientvit\_b1.r224\_in1k}                        \\
MobileNetV4$_\mathit{conv-m}$\cite{qin_mobilenetv4_2024}           & 79.07\%          & 0.8           & 41.4             & \texttt{mobilenetv4\_conv\_medium.e500\_r224\_in1k}           \\
\textbf{ORXE}                                                      & \textbf{78.95\%} & \textbf{0.4}  & \textbf{257.7}   & \textbf{$\lambda=0.200$}                                  \\
EfficientNet$_\mathit{b3}$\cite{tan_efficientnet_2019-1}           & 78.64\%          & 1.0           & 733.0            & \texttt{efficientnet\_b3.ra2\_in1k}                         \\
ResNet$_\mathit{152}$\cite{he_deep_2016}                           & 78.24\%          & 11.5          & 464.9            & \texttt{resnet152.tv\_in1k}                                \\
EfficientNet$_\mathit{b2}$\cite{tan_efficientnet_2019-1}           & 77.89\%          & 0.7           & 187.7            & \texttt{efficientnet\_b2.ra\_in1k}                          \\
{\ul EfficientNet$_\mathit{b0}$\cite{tan_efficientnet_2019-1}}     & 77.71\%          & 0.4           & 187.7            & \texttt{efficientnet\_b0.ra\_in1k}                          \\
EfficientNet$_\mathit{b1}$\cite{tan_efficientnet_2019-1}           & 77.57\%          & 0.6           & 187.7            & \texttt{efficientnet\_b1.ft\_in1k}                          \\
ResNet$_\mathit{101}$\cite{he_deep_2016}                           & 77.26\%          & 7.8           & 333.6            & \texttt{resnet101.tv\_in1k}                                \\
EfficientViT$_\mathit{m5}$\cite{liu_efficientvit_2023}             & 77.08\%          & 0.6           & 75.9             & \texttt{efficientvit\_m5.r224\_in1k}                        \\
EfficientFormerV2$_\mathit{s0}$\cite{li_rethinking_2023}           & 76.25\%          & 0.4           & 342.1            & \texttt{efficientformerv2\_s0.snap\_dist\_in1k}              \\
ResNet$_\mathit{50}$\cite{he_deep_2016}                            & 75.86\%          & 4.1           & 352.1            & \texttt{resnet50.tv\_in1k}                                 \\
{\ul MobileNetV3$_\mathit{large100}$\cite{howard_searching_2019}} & 75.78\%          & 0.2           & 70.0             & \texttt{mobilenetv3\_large\_100.ra\_in1k}                    \\ \hline
\textbf{ORXE}                                                      & \textbf{75.77\%} & \textbf{0.2}  & \textbf{115.3}   & \textbf{$\lambda=0.054$}                                  \\
EfficientViT$_\mathit{m4}$\cite{liu_efficientvit_2023}             & 74.33\%          & 0.3           & 75.9             & \texttt{efficientvit\_m4.r224\_in1k}                        \\
{\ul MobileNetV4$_\mathit{conv-s}$\cite{qin_mobilenetv4_2024}}     & 73.75\%          & 0.2           & 45.4             & \texttt{mobilenetv4\_conv\_small.e2400\_r224\_in1k}           \\
EfficientViT$_\mathit{m3}$\cite{liu_efficientvit_2023}             & 73.39\%          & 0.3           & 75.9             & \texttt{efficientvit\_m3.r224\_in1k}                        \\
ResNet$_\mathit{34}$\cite{he_deep_2016}                            & 73.20\%          & 3.7           & 352.1            & \texttt{resnet34.tv\_in1k}                                 \\
{\ul EfficientViT$_\mathit{b0}$\cite{cai_efficientvit_2023}}       & 71.36\%          & 0.1           & 147.0            & \texttt{efficientvit\_b0.r224\_in1k}                        \\
EfficientViT$_\mathit{m2}$\cite{liu_efficientvit_2023}             & 70.79\%          & 0.2           & 75.9             & \texttt{efficientvit\_m2.r224\_in1k}                        \\
ResNet$_\mathit{18}$\cite{he_deep_2016}                            & 69.54\%          & 1.8           & 352.1            & \texttt{resnet18.tv\_in1k}                                 \\
EfficientViT$_\mathit{m1}$\cite{liu_efficientvit_2023}             & 68.32\%          & 0.2           & 75.9             & \texttt{efficientvit\_m1.r224\_in1k}                        \\
{\ul MobileNetV3$_\mathit{small100}$\cite{howard_searching_2019}} & 67.64\%          & 0.1           & 34.4             & \texttt{mobilenetv3\_small\_100.lamb\_in1k}                  \\
EfficientViT$_\mathit{m0}$\cite{liu_efficientvit_2023}             & 63.27\%          & 0.1           & 75.9             & \texttt{efficientvit\_m0.r224\_in1k}                        \\ \hline
\end{tabular}%
}
\end{table*}

\section{Speed Test on More Devices}
\label{sec:additional_experiments}
\begin{figure}[htbp]
  \centering
  \begin{subfigure}{0.45\linewidth}
    \includegraphics[width=\linewidth]{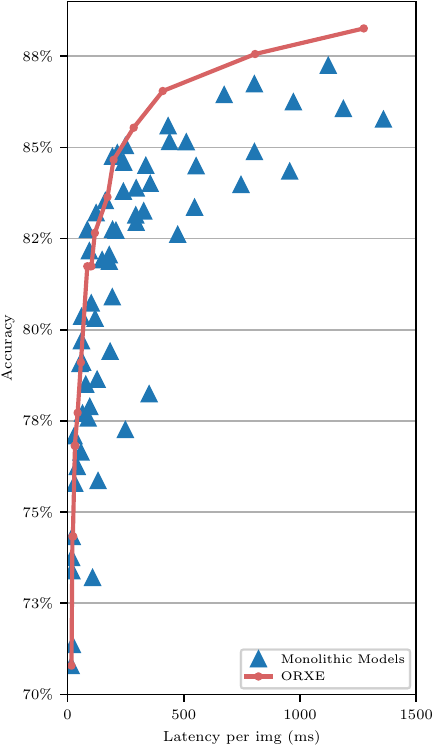}
    \caption{Raspberry Pi, BS 1}
    \label{fig:rapi1}
  \end{subfigure}
  \begin{subfigure}{0.45\linewidth}
    \includegraphics[width=\linewidth]{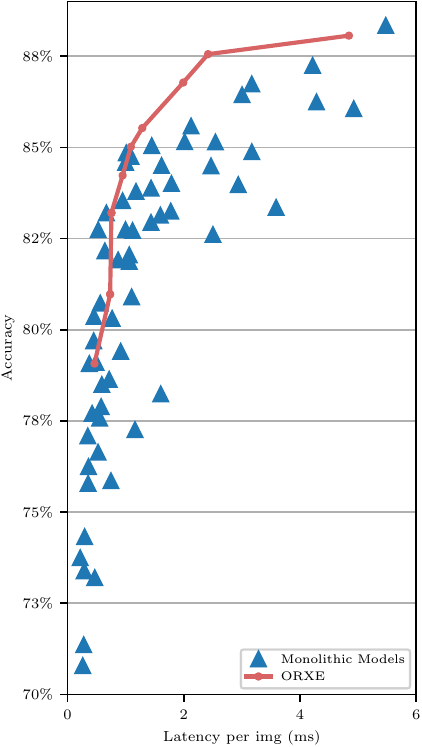}
    \caption{RTX 3090, BS 128}
    \label{fig:gpuonnx128}
  \end{subfigure}
    \caption{The speed test of ORXE and other models on Raspberry Pi and GPU}
    \label{fig:speed_test2}
\end{figure}

As a supplement to \cref{sec:on_device_test}, we further conducted speed tests on more practical scenarios. ORXE can be dynamically configurable while keeping better efficiency in most cases. \cref{fig:rapi1} shows the results on Raspberry Pi which is a small edge device. The inference latency of the high performance model is reduced by approx. 30\% to 50\% with the ORXE metamodel. \cref{fig:gpuonnx128} shows the results on RTX 3090 with a batch size of 128. The batched input results in dynamic batch sizes for the ORXE experts. Modern GPUs support batched processing, allowing the input to utilize more computing units. The proposed method still works in this scenario and have good reduction on the high accuracy model. 

\section{Efficiency in Exiting with Predicted Probability}
\begin{figure}[htbp]
  \centering
  \begin{subfigure}{\linewidth}
    \includegraphics[width=\linewidth]{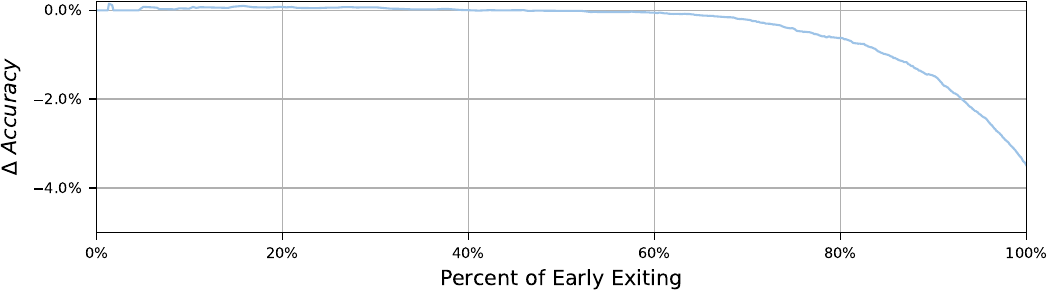}
    \caption{TinyViT$_\mathsf{21m}$\cite{wu_tinyvit_2022} - BeiT$_\mathsf{large}$\cite{peng_beit_2022}}
    \label{fig:gating_error_1}
  \end{subfigure}
  \\
  \begin{subfigure}{\linewidth}
    \includegraphics[width=1\textwidth]{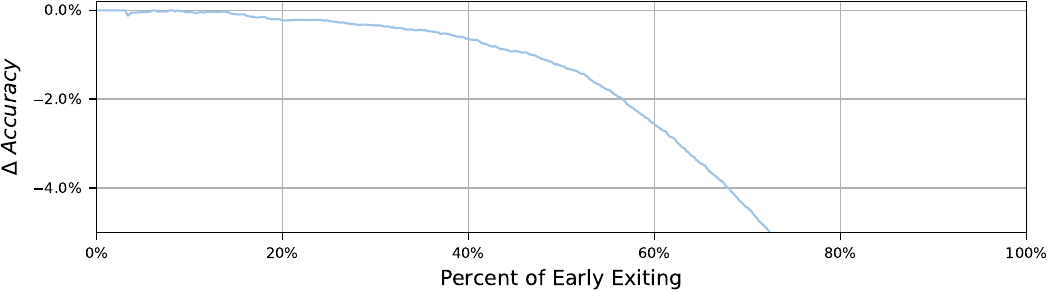}
	\caption{MobileNetV4$_\mathsf{conv-m}$\cite{qin_mobilenetv4_2024} - BeiT$_\mathsf{large}$\cite{peng_beit_2022}}
	\label{fig:gating_error_2}
  \end{subfigure}
  \caption{Accuracy difference on early exited samples}
	\label{fig:gating_error}

\end{figure}

The efficiency gains from orchestrating multiple models depend on the accuracy of the gating mechanism. A potential concern here is the unreliability of model confidence scores. Specifically, a sample that could have been correctly predicted by a larger model might be early exited due to a false positive confidence prediction by a smaller model, thus leading to errors. However, such situations do not occur frequently, especially at high confidence ranges. \cref{fig:gating_error} illustrates the difference in performance for early-exited samples between smaller and larger models. As shown in \cref{fig:gating_error_1}, a smaller model can safely rely on its confidence to select over 50\% of the samples without any performance loss compared to the larger model. In other words, the larger model does not provide significantly better predictions for these samples. A more extreme example is combining a very small model with a much larger one, where their FLOPs have 300x differences. Even in this scenario, the smaller model can still effectively handle 40\% of the workload, introducing only about 1\% additional error. Therefore, gating errors do not impede the overall efficiency gains achievable by the proposed system.
\section{Training Confidence Estimators}
\label{sec:trained_conf}
For classification models, the output probability naturally serves as an indicator of sample-level confidence. However, many models designed for other tasks do not inherently provide confidence estimates for individual samples. For instance, regression models typically output numerical predictions without accompanying confidence scores. To enable early exiting in models that do not natively provide confidence estimates, an additional evaluator is required. Such an evaluator can be trained to approximate sample-level metrics. For example, in regression tasks, the evaluator could be trained to predict the absolute error of model outputs. Nevertheless, some metrics cannot be directly obtained at the level of individual inputs. An illustrative case is the mean average precision (mAP) metric used in object detection, which is defined over an entire dataset rather than individual samples. Furthermore, these metrics often exhibit non-uniform distributions, and label imbalance can lead to overfitting of the evaluator.\\
However, confidence measures used for early exiting do not necessarily need to be strictly calibrated against specific performance metrics. The primary goal of the early exit routing is to distinguish samples by the difficulty. We hope to divide the dataset by the score into two parts where the prediction for one subset is better than the other one. Thus, samples with more reliable predictions can exit from further computation. Therefore, the confidence function $\mathit{Conf}(\cdot)$ should ideally have:
{\small
\begin{align}  
\forall x_1, x_2,\quad \begin{split}&\bigl(\mathit{Conf}(x_1)>th \land \mathit{Conf}(x_2)\le th\bigr) \\ &\implies \mathit{Metr}(x_1)>\mathit{Metr}(x_2)\end{split}
\end{align}
}
Moreover, the threshold value should be arbitrary while configuring the overall cost and performance. Hence, the actual target of $\mathit{Conf}(\cdot)$ in a configurable early exiting system is ideally to obtain the property of:
{\small
\begin{align}
\forall x_1, x_2,\quad \begin{split} &\mathit{Conf}(x_1)>\mathit{Conf}(x_2)\\ &\implies \mathit{Metr}(x_1)>\mathit{Metr}(x_2)\end{split} \label{eq:conf_target}
\end{align}
}
Where $\mathit{Metr}(x)$ represents the actual metric value of model's prediction on $x$ and $\mathit{Conf}(x)$ denotes the estimated confidence value of that prediction. \\
Therefore, the target of $Conf(\cdot)$, which is expressed in \cref{eq:conf_target}, is ranking samples by $\mathit{Metr}(x_1)$ rather than fitting to it. Now the metric can be any value indicating the relative prediction reliability of a sample, like the negative loss value which is available in every supervised learning task.

\begin{figure}[htbp]
    \centering
    \includegraphics[width=0.6\linewidth]{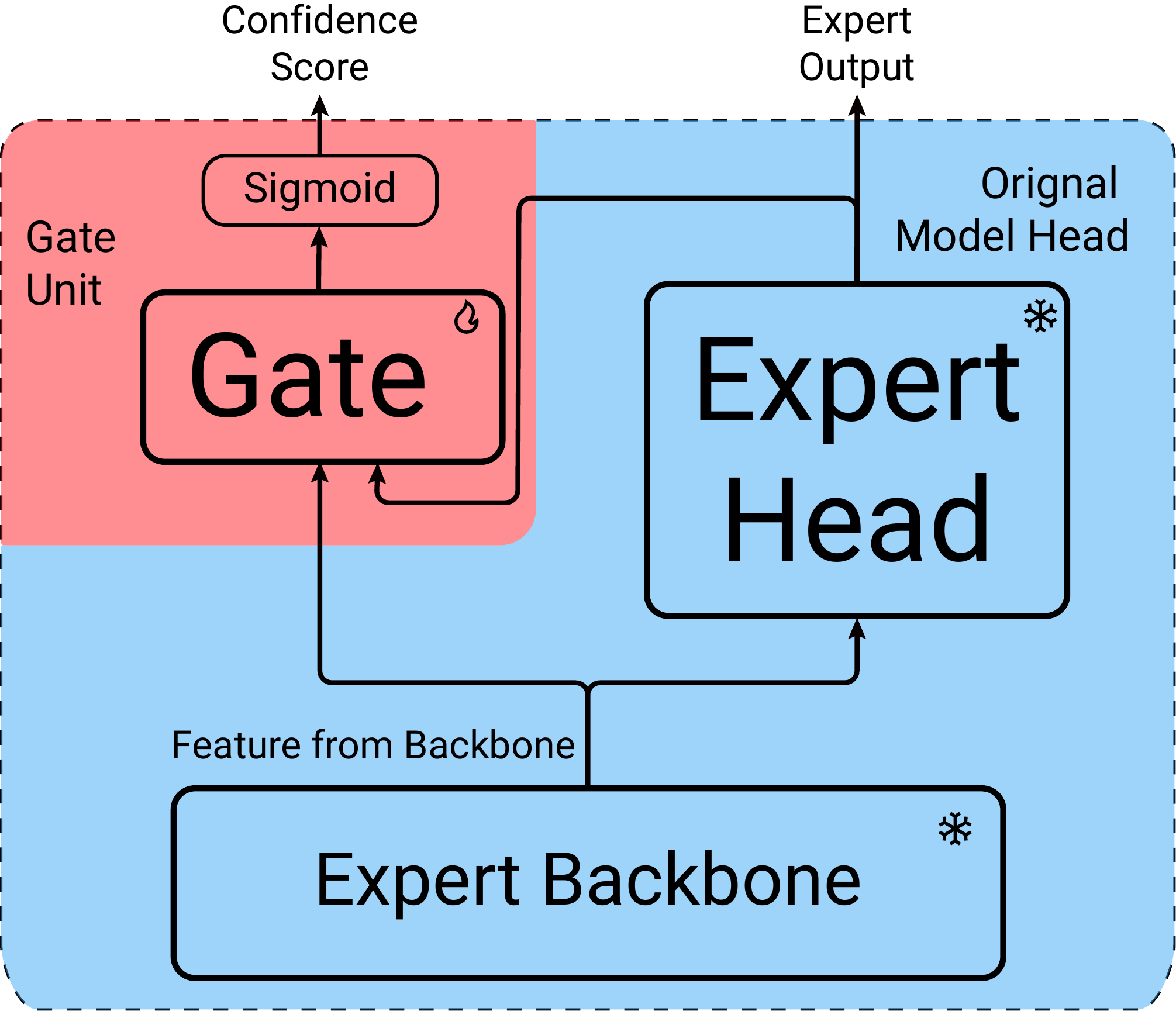}
    \caption{The gate architecture}
    \label{fig:gate_architecture}
\end{figure}

\paragraph{Ranking as Calibration}
Pairwise comparison is a simple but effective method to train neural networks to relative relationship. It asks models to compare arbitrary two samples and learn to predict which one should rank higher. With comprehensive training, the model will be able to produce appropriate score to every sample with a sorted ranking. Concretely, we form $\frac{B(B-1)}{2}$ pairs for a batch with $B$ samples. As described in \cref{eq:pairwise_target}, each pair between sample predictions $i$ and $j$ is assigned with a label to denote which prediction is better.
{\small
\begin{align}
\mathit{target}_{i,j} & = \begin{cases}
1  & \text{ if } \mathit{Metr}(x_i)>\mathit{Metr}(x_j) \\
0.5  & \text{ if } \mathit{Metr}(x_i) = \mathit{Metr}(x_j)\\
 0 & \text{ if } \mathit{Metr}(x_i)<\mathit{Metr}(x_j)
\end{cases}\label{eq:pairwise_target}
\end{align}
}
Then the binary cross entropy loss is employed to guide the model to match the relative order between the predicted logit and the metric within a pair. Such a loss function is only sensitive to the relative relationship but not the absolute value of the metric. It simplifies the design of metric value and also resolves the label imbalance issue.
{\small
\begin{align}
\ell & = \frac{2}{B(B-1)}\sum_{i  = 1}^{B} \sum_{j  = i}^{B} \mathit{BCE}(\mathit{Gate}(x_i)-\mathit{Gate}(x_j),\mathit{target}_{i,j}) \label{eq:gate_loss}
\end{align}
}
Finally, the used confidence value is derived from the gate output with a sigmoid transformation. The sigmoid function does not change the relative order of its input, but limiting the output to $[0,1]$ for the convenience of threshold searching.
{\small
\begin{align}
\mathit{Conf}(x) & = \mathit{Sigmoid}(\mathit{Gate}(x))
\end{align}
}
\paragraph{Gates Architecture}
The purpose of early exiting is to save computation. Therefore, the gate must be implemented in a highly simple form to reduce the overhead to the system. In this work, the gate is designed to be small MLPs. As depicted in \cref{fig:gate_architecture}, the gate receives inputs from the model backbone and the output. The feature from the backbone is pooled for reducing its shape. The model output, subject to the specific model, may need an embedding layer before entering the MLP. Please refer the experiment section for the specific design of the gate. The output of the MLP is a logit value $\mathit{Gate}(x)$ which is trained to express the ranking for the prediction of the attached expert model on the sample $x$. 

\section{Relation to EE and MoE}
The proposed ORXE model can be viewed as a special case of early exiting (EE) and top-1 mixture of experts (MoE). However, there are fundamental differences between our work and conventional EE and MoE.\\
In conventional EE systems, the model is partitioned into multiple sequential segments, each augmented with an auxiliary head for intermediate predictions. Computation must always begin from the first segment, deciding at each head whether to proceed or stop further computation. In contrast, within ORXE, each segment functions as an independent expert model; therefore, computation is not constrained to always start from the initial segment. During routing, ORXE not only supports the standard EE decision-making (continue or exit) but also enables very challenging samples to bypass multiple intermediate experts directly. Consequently, ORXE provides significantly greater flexibility compared to conventional EE methods. The comparative results in \cref{fig:compare_with_dynamic} further demonstrate that ORXE achieves higher efficiency than conventional EE approaches.
\begin{figure*}[htbp]
  \centering
  \begin{subfigure}[b]{0.45\linewidth}
  \centering
    \includegraphics[width=0.45\linewidth]{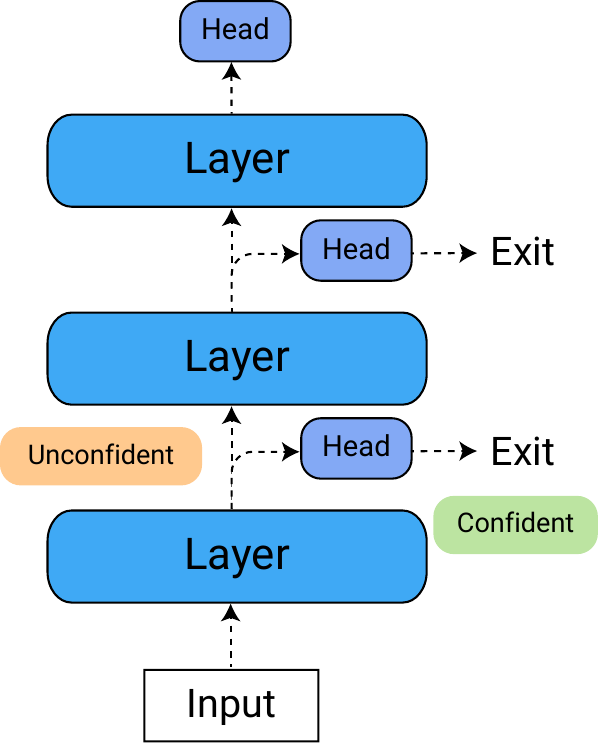}
    \caption{Conventional Early Exit}
    \label{fig:ee:normal_ee}
  \end{subfigure}
  \begin{subfigure}[b]{0.5\linewidth}
  \centering
    \includegraphics[width=0.5\linewidth]{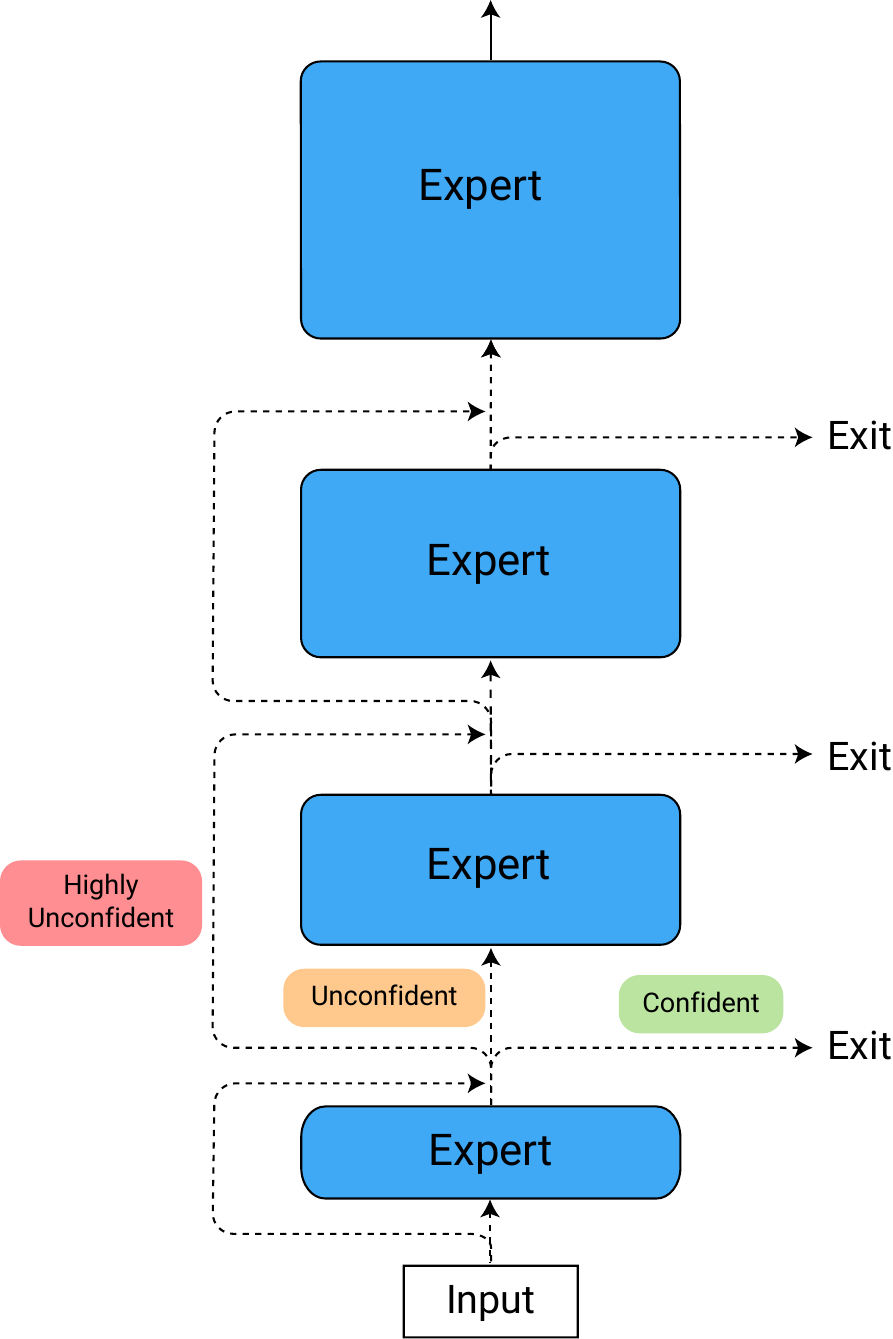}
	\caption{ORXE's equivalence in the context of Early Exit}
	\label{fig:ee:orxe_ee}
  \end{subfigure}
  \caption{Differences between ORXE and conventional EE}
	\label{fig:ee}

\end{figure*}
Within a MoE framework, the primary advantage of ORXE lies in leveraging a subset of experts themselves to determine the data routing path rather than a dedicated router. In conventional Top-1 MoE systems, the routing decision is made prior to invoking any experts. Consequently, the router must possess sufficient complexity and capacity to accurately understand the relative strengths of various experts. Such a router is computationally costly, especially in systems designed explicitly for computational efficiency. In contrast, ORXE adopts an approach where experts are selectively used on-demand as implicit routers. This design eliminates the additional computational overhead associated with dedicated routing modules. Furthermore, ORXE can be viewed as employing a form of partial post-expert routing, inherently providing greater accuracy compared to purely pre-expert routing methods.

\begin{figure*}[htbp]
  \centering
  \begin{subfigure}[b]{0.5\linewidth}
    \centering
    \includegraphics[width=0.5\linewidth]{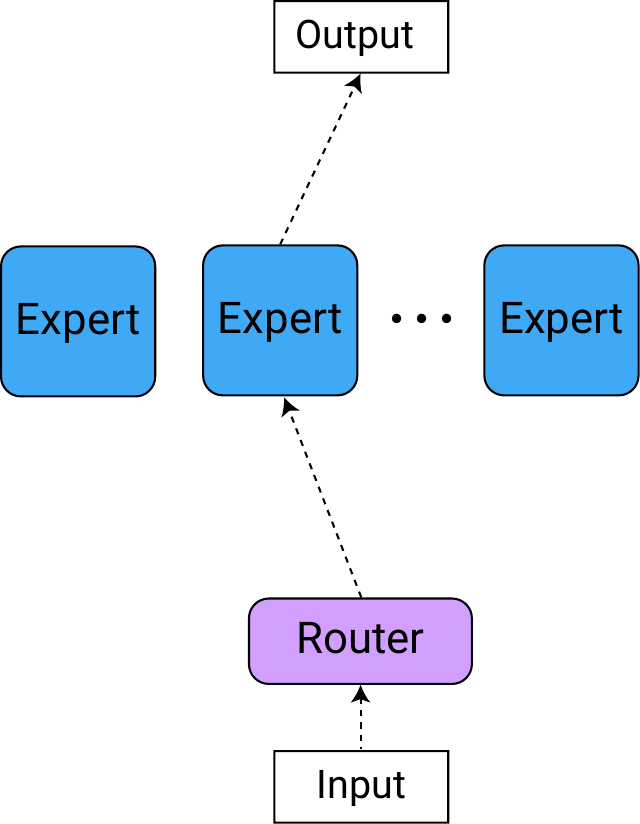}
    \caption{Conventional Top-1 Mixture of Experts}
    \label{fig:moe:normal_moe}
  \end{subfigure}%
  \begin{subfigure}[b]{0.5\linewidth}
    \centering
    \includegraphics[width=0.5\linewidth]{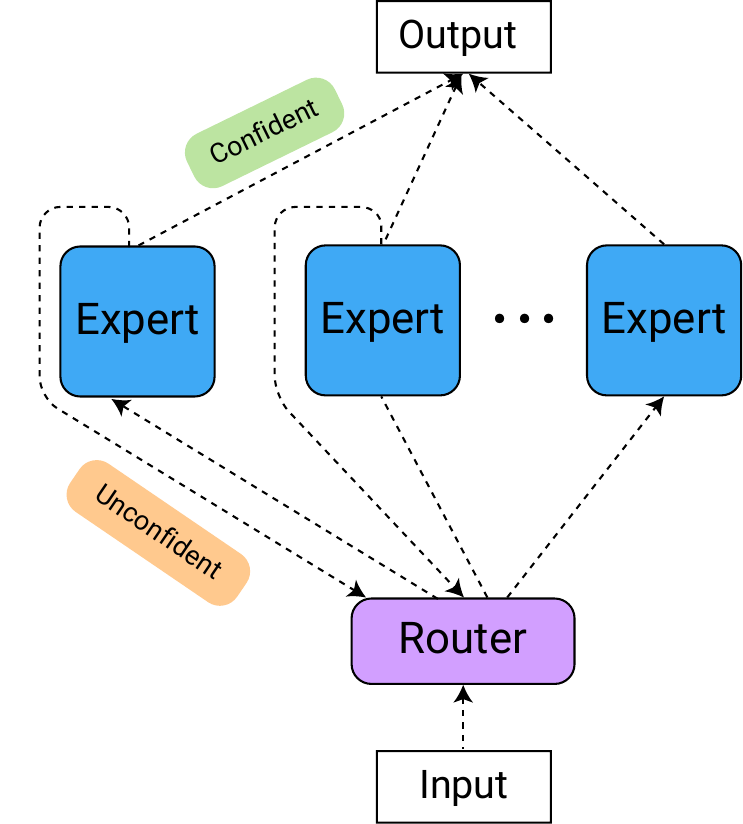}
    \caption{ORXE's equivalence in the context of Mixture of Experts}
    \label{fig:moe:orxe_moe}
  \end{subfigure}
  \caption{Differences between ORXE and conventional Top-1 MoE}
  \label{fig:moe}
\end{figure*}